\documentclass[10pt,twocolumn,letterpaper]{article}
\usepackage{iccv}
\usepackage{times}
\usepackage{epsfig} 
\usepackage{comment}  
\usepackage{multirow}  
\usepackage{booktabs}
\usepackage{varwidth}

\usepackage[ruled,vlined]{algorithm2e}    
\usepackage{stackengine} 
\usepackage{bm}
\usepackage{color}
\usepackage{empheq}

\usepackage{microtype} 
\usepackage{graphicx}  
\usepackage{amsmath,amssymb}
\usepackage[breaklinks=true,colorlinks,bookmarks=false]{hyperref} 
\usepackage[breaklinks=true,bookmarks=false]{hyperref}

\DeclareMathOperator*{\argmax}{arg\,max}

\iccvfinalcopy 
\def\iccvPaperID{****} 

\ificcvfinal\fi

\begin{document}
%%%%%%%%% TITLE 
\title{Mixture-based Feature Space Learning for Few-shot Image Classification} 
\author{Arman Afrasiyabi$^{\star \bullet}$, %\inst{1}\orcidID{0000-0002-7346-3319} \and
Jean-Fran\c{c}ois Lalonde$^\star$, % \inst{2,3}\orcidID{0000-0002-6583-2364} \\ \and
Christian Gagn\'e$^{\star \dag \bullet}$ %\inst{3}\orcidID{0000-0003-3697-4184}
\\
$^\star$Universit\'e Laval, $^\dag$Canada CIFAR AI Chair, $^\bullet$Mila   \\ 
\small{\texttt{\url{https://lvsn.github.io/MixtFSL/}}} 
} 
\newcommand\myparagraph[1]{\vspace{0.25em}\noindent\textbf{#1}\quad}
\maketitle
% Remove page # from the first page of camera-ready.
% \ificcvfinal \thispagestyle{empty}\fi

\begin{abstract}
We introduce Mixture-based Feature Space Learning (MixtFSL) for obtaining a rich and robust feature representation in the context of few-shot image classification. Previous works have proposed to model each base class either with a single point or with a mixture model by relying on offline clustering algorithms. In contrast, we propose to model base classes with mixture models by simultaneously training the feature extractor \emph{and} learning the mixture model parameters in an online manner. This results in a richer and more discriminative feature space which can be employed to classify novel examples from very few samples. Two main stages are proposed to train the MixtFSL model. First, the multimodal mixtures for each base class and the feature extractor parameters are learned using a combination of two loss functions. Second, the resulting network and mixture models are progressively refined through a leader-follower learning procedure, which uses the current estimate as a ``target'' network. This target network is used to make a consistent assignment of instances to mixture components, which increases performance and stabilizes training. The effectiveness of our end-to-end feature space learning approach is demonstrated with extensive experiments on four standard datasets and four backbones. Notably, we demonstrate that when we combine our robust representation with recent alignment-based approaches, we achieve new state-of-the-art results in the inductive setting, with an absolute accuracy for 5-shot classification of 82.45\% on miniImageNet, 88.20\% with tieredImageNet, and 60.70\% in FC100 using the ResNet-12 backbone.
\end{abstract}

\section{Introduction}
\label{sec:introduction}

%$ paragraph-1: the few-shot problem
The goal of few-shot image classification is to transfer knowledge gained on a set of ``base'' categories, containing a large number of training examples, to a set of distinct ``novel'' classes having very few examples~\cite{fei2006one,mnih2015human}. 
A hallmark of successful approaches~\cite{finn2017model, snell2017prototypical, vinyals2016matching} is their ability to learn rich and robust feature \emph{representations} from base training images, which can generalize to novel samples.

A common assumption in few-shot learning is that classes can be represented %(in feature space) 
with unimodal models. For example, Prototypical Networks~\cite{snell2017prototypical} (``ProtoNet'' henceforth) assumed each base class can be represented with a single prototype. Others, favoring standard transfer learning~\cite{Afrasiyabi_2020_ECCV,chen2019closer,gidaris2018dynamic}, use a classification layer which push each training sample towards a single vector. While this strategy has successfully been employed in ``typical'' image classification (\eg, ImageNet challenge~\cite{russakovsky2015imagenet}), it is somewhat counterbalanced because the learner is regularized by using validation examples that belong to the same training classes. Alas, this solution does not transfer to few-shot classification since the base, validation, and novel classes are disjoint. Indeed, Allen~\etal~\cite{allen2019infinite} showed that relying on that unimodal assumption limits adaptability in few-shot image classification and is prone to underfitting from a data representation perspective.

%% paragraph-3: IMP and why mixture model is good   
To alleviate this limitation, Infinite Mixture Prototypes~\cite{allen2019infinite} (IMP) extends ProtoNet by representing each class with multiple centroids.
This is accomplished by employing an offline clustering (extension of DP-means~\cite{kulis2012revisiting}) where the non-learnable centroids are recomputed at each iteration. % of the training. 
This approach however suffers from two main downsides. First, it does not allow capturing the global distribution of base classes since a small subset of the base samples are clustered at any one time---clustering over all base samples at each training iteration would be prohibitively expensive. 
Second, relying on DP-means in an offline, post hoc manner implies that feature learning and clustering are done independently. % of each other. 

In this paper, we propose ``Mixture-based Feature Space Learning'' (MixtFSL) to learn a multimodal representation for the base classes 
using a mixture of trainable components---learned vectors that are iteratively refined during training. The key idea is to learn both the \emph{representation} (feature space) and the \emph{mixture model} jointly in an online manner, which effectively unites these two tasks by allowing the gradient to flow between them. This results in a discriminative  representation, which in turn yields superior performance when training on the novel classes from few examples.

\begin{figure} 
    \centering
    \footnotesize
    \setlength{\tabcolsep}{1pt}
    \begin{tabular}{ccc} 
    \includegraphics[width=0.45\linewidth, angle=0]{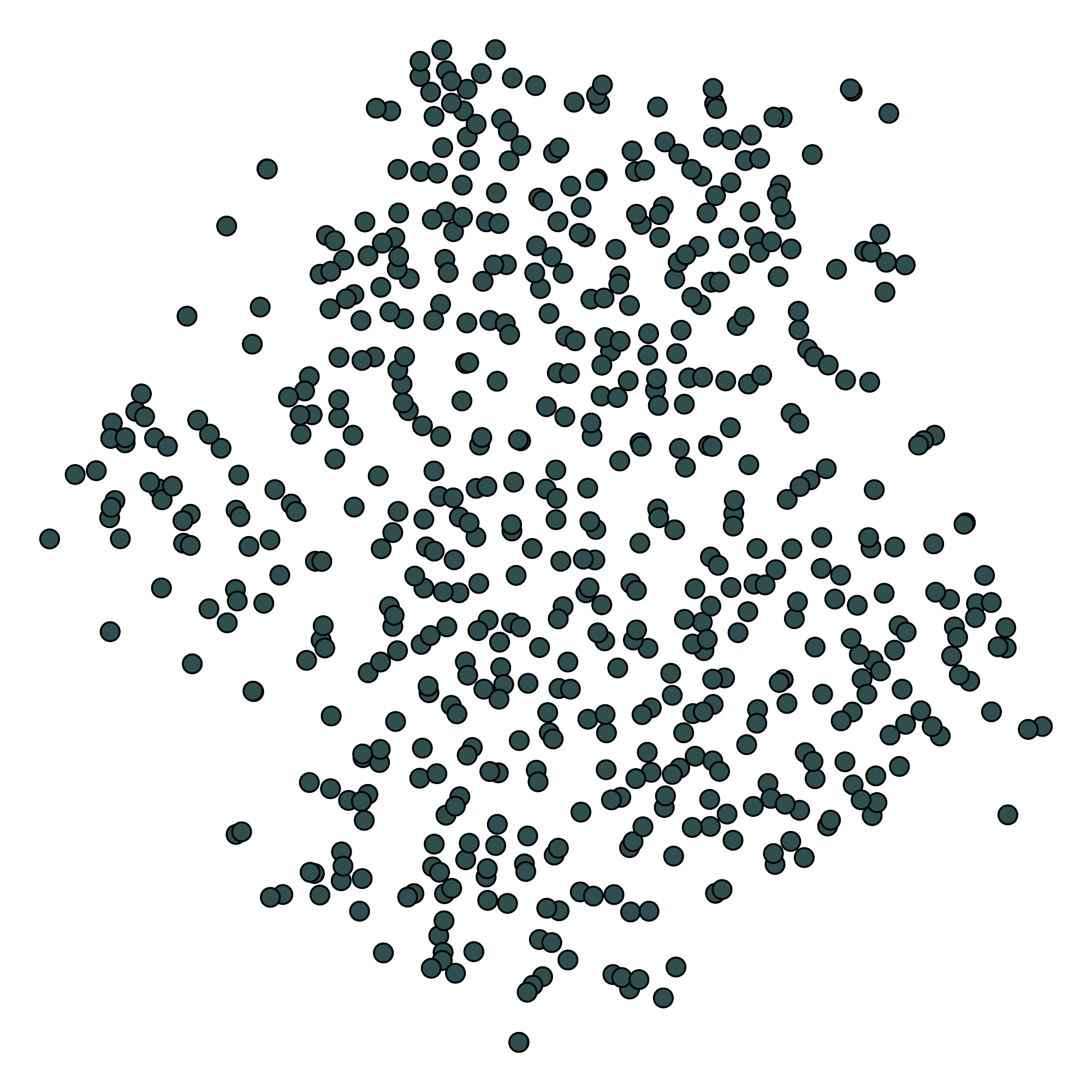} & 
    \includegraphics[width=0.47\linewidth, angle=0]{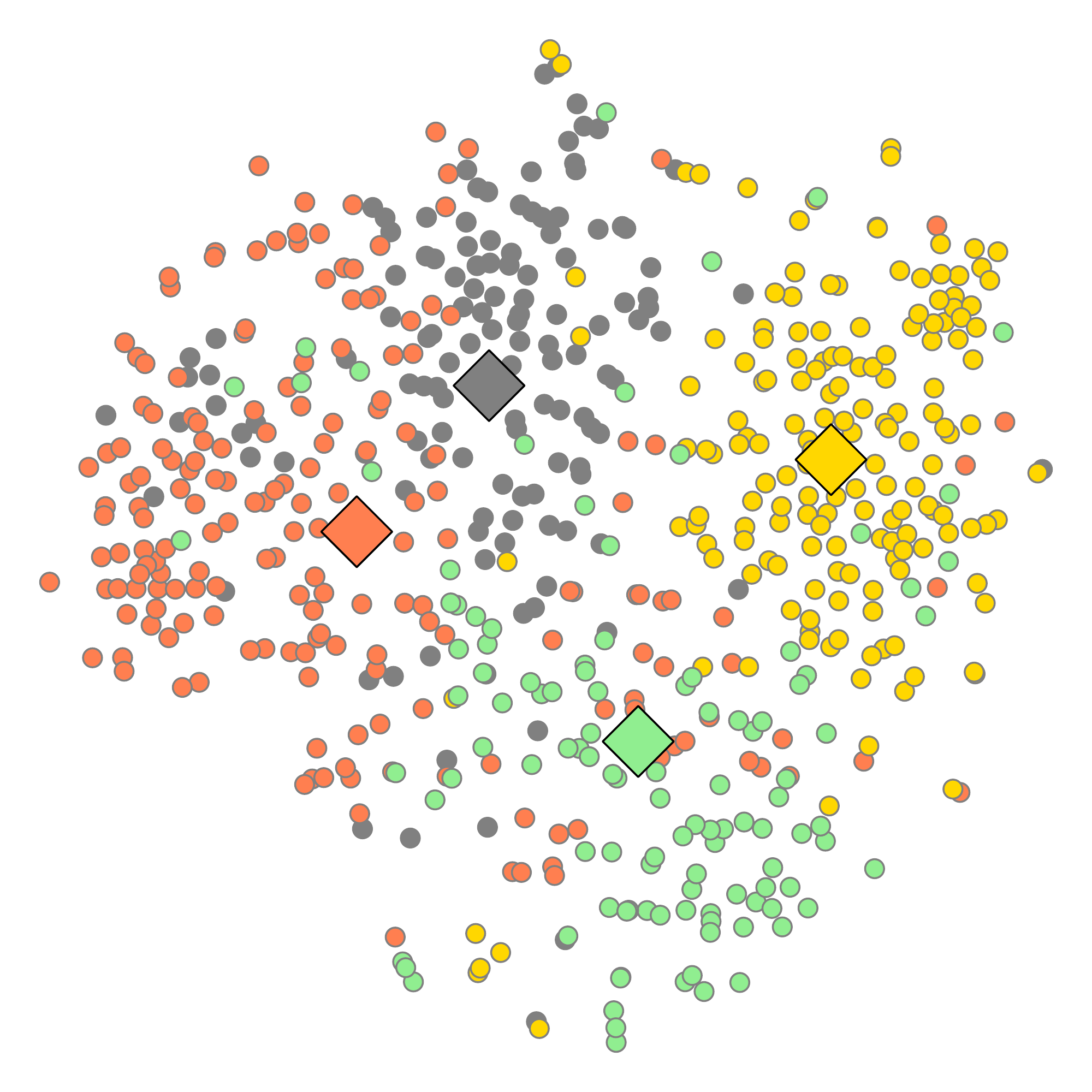} &   \\ 
    (a) without MixtFSL  & 
    (b) our MixtFSL \\       
    \end{tabular}
    \vspace{0.5em}
    \caption{t-SNE~\cite{maaten2008visualizing} visualization of a single base class embedding (circles) (a) without, and (b) with our MixtFSL approach. MixtFSL learns a representation for base samples (circles) and associated mixture learned components (diamonds) that clusters a class into several modes (different colors). This more flexible representation helps in training robust classifiers from few samples in the novel domain compared to the monolithic representation of (a). 
    Embeddings are extracted from a miniImageNet with a ResNet-18.}
    \label{fig:idea}
\end{figure}

We propose a two-stage approach to train our MixtFSL. In the first stage, the mixture components are initialized by the combination of two losses that ensure that: 1) samples are assigned to their nearest mixture component; while 2) enforcing components of a same class mixture to be far enough from each other, to prevent them from collapsing to a single point. In the second stage, the learnable mixture model is progressively refined through a leader-follower scheme, which uses the current estimate of the learner as a fixed ``target'' network, updated only on a few occasions during that phase, and a progressively declining temperature strategy. Our experiments demonstrate that this improves performance and stabilizes the training. During training, the number of components in the learned mixture model is automatically adjusted from data. The resulting representation is flexible and  better adapts to the multi-modal nature of images (fig.~\ref{fig:idea}), which results in improved performance on the novel classes.

Our contributions are as follows. We introduce the idea of MixtFSL for few-shot image classification, which learns a flexible representation by modeling base classes as a mixture of learnable components. We present a robust two-stage scheme for training such a model. The training is done end-to-end in a fully differentiable fashion, without the need for an offline clustering method. We demonstrate, through an extensive experiments on four standard datasets and using four backbones, that our MixtFSL outperforms the state of the art in most of the cases tested. We show that our approach is flexible and can leverage other improvements in the literature (we experiment with associative alignment~\cite{Afrasiyabi_2020_ECCV} and ODE~\cite{xu2021learning}) to further boost performance. Finally, we show that our approach does not suffer from forgetting (the base classes).

\section{Related work}
\label{sec:related-work}

Few-shot learning is now applied to problems such as image-to-image translation~\cite{Wang1_2020_CVPR}, object detection~\cite{Fan1_2020_CVPR, Perez-Rua_2020_CVPR}, video classification~~\cite{Cao_2020_CVPR}, and 3D shape segmentation~\cite{Wang3_2020_CVPR}. This paper instead focuses on the image classification problem~\cite{finn2017model,snell2017prototypical,vinyals2016matching}, so the remainder of the discussion will focus on relevant works in this area. In addition, unlike transductive inference methods~\cite{boudiaf2020transductive, dhillon2019baseline, NEURIPS2019_01894d6f, Liu_2020_ECCV, kim2019edge, liu2018learning, masud2020laplacian, Qiao_2019_ICCV} which uses the structural information of the entire novel set, our research focuses on inductive inference research area. %based few-shot image classification.  

%% episodic training  
\myparagraph{Meta learning}
In meta learning~\cite{dhillon2019baseline,finn2017model,ravi2016optimization,rusu2018meta,simon2020modulating,snell2017prototypical,su2020does,vilalta2002perspective,wang2019meta,yoon2019tapnet}, approaches imitate the few-shot scenario by repeatedly sampling similar scenarios (episodes) from the base classes during the pre-training phase. Here, distance-based approaches~\cite{bertinetto2018metalearning,garcia2017few,kim2019variational,Li_2019_CVPR,lifchitz2019dense,oreshkin2018tadam,snell2017prototypical,sung2018learning,tseng2020cross,vinyals2016matching,wertheimer2019few,Zhang_2020_CVPR,zhang2019variational} aim at transferring the reduced intra-class variation from base to novel classes, while initialization-based approaches~\cite{finn2017model,finn2018probabilistic,kim2018bayesian} are designed to carry the best starting model configuration for novel class training. Our MixtFSL benefits from the best of both worlds, by reducing the within-class distance with the learnable mixture component and increasing the adaptivity of the network obtained after initial training by representing each class with mixture components.

%% standard transfer learning 
\myparagraph{Standard transfer learning} 
Batch form training makes use of a standard transfer learning \emph{modus operandi} instead of episodic training.   
Although batch learning with a naive optimization criteria is prone to overfitting, several recent studies~\cite{Afrasiyabi_2020_ECCV, chen2019closer, gidaris2018dynamic,Qi_2018_CVPR, Tian_2020_ECCV_good} have shown a metric-learning (margin-based) criteria can offer good performance. For example, Bin et al.~\cite{Bin_2020_ECCV_margin_matter} present a negative margin based feature space learning.  
Our proposed MixtFSL also uses transfer learning but innovates by simultaneously clustering base class features into multi-modal mixtures in an online manner.

\myparagraph{Data augmentation}
Data augmentation~\cite{Chen_2019_CVPR,Chu_2019_CVPR,gao2018low,gidaris2019boosting,gidaris2019generating,hariharan2017low,Liu_2019_ICCV,mehrotra2017generative,ren2018meta,schwartz2018delta,wang2018low,wang2016learning,zhang2017mixup,Zhang_2019_CVPR, zhang2020iept} for few-shot image classification aims at training a well-generalized algorithm. Here, the data can be augmented using a generator function. For example, \cite{hariharan2017low} proposed Feature Hallucination (FH) using an auxiliary generator. Later, \cite{wang2018low} extends FH to generate new data using generative models. In contrast, our MixtFSL does not generate any data and achieves state-of-the-art. \cite{Afrasiyabi_2020_ECCV} makes use of ``related base'' samples in combination with an alignment technique to improve performance. We demonstrate (in sec.~\ref{sec:extensions}) that we can leverage this approach in our framework since our contribution is orthogonal.

\myparagraph{Mixture modeling}
Similar to classical mixture-based works~\cite{fernando2012supervised,gauvain1994maximum} outside few-shot learning, infinite mixture model~\cite{hjort2010bayesian} explores Bayesian methods~\cite{rasmussen2000infinite,west1993hierarchical} to infer the number of mixture components. Recently, IMP~\cite{allen2019infinite} relies on the DP-means~\cite{kulis2012revisiting} algorithm which is computed inside the episodic training loop in few-shot learning context. As in~\cite{hjort2010bayesian}, our MixtFSL automatically learns the number of mixture components, but differs from \cite{allen2019infinite} by learning the mixture model simultaneously with representation learning in an online manner, without the need for a separate, post hoc clustering algorithm. From the learnable component perspective, our MixtFSL is related to VQ-VAE~\cite{NEURIPS2019_5f8e2fa1,van2017neural} which learns quantized feature vectors for image generation, and SwAV~\cite{caron2020unsupervised} for self-supervised learning. 
Here, we tackle supervised few-shot learning by using mixture modeling to increase the adaptivity of the learned representation. This also contrasts with variational few-shot learning~\cite{kim2019variational,zhang2019variational}, which aims to reduce noise with variational estimates of the distribution. Our MixtFSL is also related to MM-Net~\cite{Cai_2018_CVPR} in that they both works store information during training. Unlike MM-Net, which contains read/write controllers plus a contextual learner to build an attention-based inference, our MixtFSL aims at modeling the multi-modality of the base classes with only a set of learned components.

% and SupCon~\cite{NEURIPS2020_d89a66c7}  

\section{Problem definition}
\label{sec:problemDefinition}

\begin{figure}
    \centering
    \includegraphics[width=\linewidth]{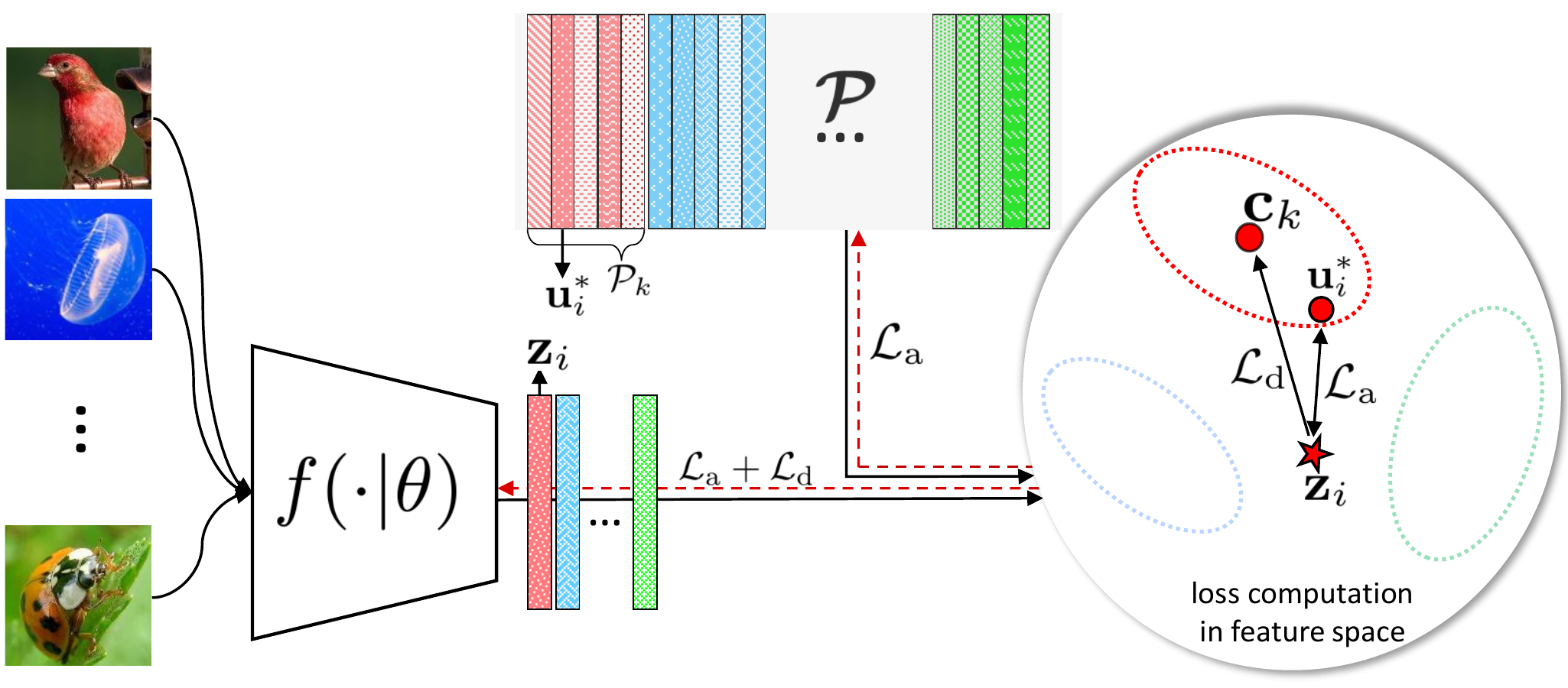}
    \caption{Initial training stage. The network $f(\cdot|\theta)$ embeds a batch (left) from the base classes to feature space. A feature vector $\mathbf{z}_i$ (middle) belonging to the $k$-th class 
    is assigned to the most similar component $\mathbf{u}^*_i$ in class mixture $\mathcal{P}_k \in \bm{\mathcal{P}}$. Vectors are color-coded by class.
    Here, two losses interact for representation learning: $\mathcal{L}_\mathrm{a}$ which maximizes the similarity between $\mathbf{z}_i$ and $\mathbf{u}^*_i$; and $\mathcal{L}_\mathrm{d}$ keeps $\mathbf{z}_i$ close to the centroid $\mathbf{c}_k$ of all mixture components for class $k$. 
    The backpropagated gradient is shown with red dashed lines. While $f(\cdot|\theta)$ is updated by $\mathcal{L}_\mathrm{it}$ (eq.~\ref{eq:loss-it}), $\bm{\mathcal{P}}$ is updated by $\mathcal{L}_\mathrm{a}$ only to prevent collapsing of the components in $\mathcal{P}_k$ to a single point.   
    }
    \label{fig:architecture}
\end{figure}

In few-shot image classification, we assume there exists a ``base'' set $\mathcal{X}^b=\{(\mathbf{x}_i,y_i)\}_{i=1}^{N^b}$, where $\mathbf{x}_i\in\mathbb{R}^D$ and $y_i\in\mathcal{Y}^b$ are respectively the $i$-th input image and its corresponding class label. There is also a ``novel'' set $\mathcal{X}^n=\{(\mathbf{x}_i,y_i)\}_{i=1}^{N^n}$, where $y_i\in\mathcal{Y}^n$, and a ``validation'' set $\mathcal{X}^v=\{(\mathbf{x}_i,y_i)\}_{i=1}^{N^v}$, where $y_i\in\mathcal{Y}^v$. None of these sets overlap and $N^n \ll N^b$.

In this paper, we follow the standard transfer learning training strategy (as in, for example, \cite{Afrasiyabi_2020_ECCV,chen2019closer}). A network $\mathbf{z} = f(\mathbf{x}|\theta)$, parameterized by $\theta$, is pre-trained to project input image $\mathbf{x}$ to a feature vector $\mathbf{z} \in \mathbb{R}^M$ using the base categories $\mathcal{X}^b$, validated on $\mathcal{X}^v$.  
%to map input $\mathbf{x} \in \mathbb{R}^D$ to feature vector $\mathbf{z} $. 
% 
The key idea behind our proposed MixtFSL model is to simultaneously train a learnable mixture model, along with $f(\cdot|\theta)$, in order to capture the distribution of each base class in $\mathcal{X}^b$. This mixture is guiding the representation learning for a better handling of multimodal class distributions, while allowing to extract information on the base class components that can be useful to stabilize the training.
We denote the mixture model across all base classes as the set $\bm{\mathcal{P}} = \{(\mathcal{P}_k, y_k)\}_{k=1}^{N^b}$, where each $\mathcal{P}_k=\{\mathbf{u}_j\}_{j=1}^{N^k}$ is the set of all $N^k$ components $\mathbf{u}_j \in \mathbb{R}^M$ assigned to the $k$-th base class. 
After training on the base categories, fine-tuning the classifier on the \emph{novel} samples is very simple and follows~\cite{chen2019closer}: the weights $\theta$ are fixed, and a single linear classification layer $\mathbf{W}$ is trained as in $c(\cdot|\mathbf{W})\equiv\mathbf{W}^\top f(\cdot|\theta)$, followed by softmax. The key observation is that the mixture model, trained only on the base classes, makes the learned feature space more discriminative---only a simple classification layer can thus be trained on the novel classes.

\section{Mixture-based Feature Space Learning}  
\label{sec:MixtFSL}

\begin{algorithm}[t] 
    \SetAlgoLined
    {\small
        \KwData{feature extractor $f(\cdot|{\theta})$, mixture $\bm{\mathcal{P}}$, base dataset $\mathcal{X}^b$, validation dataset $\mathcal{X}^v$, maximum epoch $\alpha_0$, patience $\alpha_1$, and error evaluation function $E(\cdot)$} %, and maximum epoch $\alpha_2$}
        \KwResult{Model $f(\cdot|{\theta^\mathrm{best}})$ and mixture $\bm{\mathcal{P}}^\mathrm{best}$ learned}
        $\theta^\mathrm{best} \leftarrow \theta$; $\bm{\mathcal{P}}^\mathrm{best} \leftarrow \bm{\mathcal{P}}$; $t \leftarrow 0$; $s \leftarrow 0$\\
        \While{$s<\alpha_0 \text{ and } t < \alpha_1$}
        { 
            \For{$(\mathbf{x}_i, y_i) \in \mathcal{X}^b$}{ 
            Evaluate $\mathbf{z}_i \leftarrow f(\mathbf{x}_i|\theta)$, and %by forward propagation\\
            %Evaluate 
            $\mathbf{u}^*_i$ %of $\mathbf{z}_i$ 
            by eq.~\ref{eq:argmax_cp} \\
            Update weights $\theta$ and mixture $\bm{\mathcal{P}}$ with $\mathcal{L}_\mathrm{it}$ (eq.~\ref{eq:loss-it}); 
            }
        Evaluate $f(\cdot|{\theta})$ on $\mathcal{X}^v$ with episodic training \\ 
        \eIf{$E(\theta,\bm{\mathcal{P}}|\mathcal{X}^v)<E(\theta^\mathrm{best},\bm{\mathcal{P}}^\mathrm{best}|\mathcal{X}^v)$}{
            $\theta^\mathrm{best}\leftarrow \theta$; $\bm{\mathcal{P}}^\mathrm{best} \leftarrow \bm{\mathcal{P}}$; $t \leftarrow 0$
        }{
            $t \leftarrow t+1$
          }
        $s \leftarrow s+1$
          %$e \leftarrow e+1$
        }
        \caption{Initial training.}
        \label{algo:initial_train}
        }
\end{algorithm}
Training our MixtFSL on the base classes is done in two main stages: initial training and progressive following. %Details on each of these stages are given below.

\subsection{Initial training}

The initial training of the feature extractor $f(\cdot|\theta)$ and the learnable mixture model $\bm{\mathcal{P}}$ from the base class set $\mathcal{X}^b$ is detailed in algorithm~\ref{algo:initial_train} and illustrated in fig.~\ref{fig:architecture}. 
In this stage, model parameters are updated using two losses: the ``assignment'' loss $\mathcal{L}_\mathrm{a}$, which updates both the feature extractor and the mixture model such that feature vectors are assigned to their nearest mixture component; and the ``diversity'' loss $\mathcal{L}_\mathrm{d}$, which updates the feature extractor to diversify the selection of components for a given class. 
Let us define the following angular margin-based softmax function~\cite{deng2018arcface}, modified with a temperature variable $\tau$: 
\begin{align}
& p_{\theta}(v_{j}|\mathbf{z}_i,\bm{\mathcal{P}}) = \label{eq:Compscoreing}\\
& \frac{e^{\cos\left( (\angle(\mathbf{z}_i,\mathbf{u}_j)+m)\right)/ \tau}}
{e^{\cos\left( (\angle(\mathbf{z}_i,\mathbf{u}_j)+m)\right) / \tau}
+\sum\limits_{\mathbf{u}_l\in\{\bm{\mathcal{P}}\backslash\mathbf{u}_{j}\}} 
e^{\cos\left( \angle(\mathbf{z}_i,\mathbf{u}_l)\right)/ \tau}} \,, \nonumber
\end{align} 
where, $m$ is a margin; $v_{j}$ is the pseudo-label  
associated to $\mathbf{u}_{j}$; and, $\angle(\mathbf{z}_i, \mathbf{u}_j)=\arccos \left(\mathbf{z}_i^\top \mathbf{u}_i/ (||\mathbf{z}_i|| ||\mathbf{u}_j||) \right)$\footnote{As per~\cite{deng2018arcface}, we 
avoid computing the arccos (which is undefined outside the $[-1,1]$ interval) and 
directly compute the $\cos(\angle(\mathbf{z}_i ,\mathbf{u}_j)+m)$.}. % (on the $L_2$-normalized vectors).}.% using simple trigonometric identities. 

Given a training image $\mathbf{x}_i$ from base class $y_i=k$ and its associated feature vector $\mathbf{z}_i = f(\mathbf{x}_i|\theta)$, the closest component $\mathbf{u}^*_i$ 
is found amongst all elements of mixture $\mathcal{P}_k$ associated to the same class according to cosine similarity:
\begin{equation} 
\label{eq:argmax_cp}
\mathbf{u}_i^* = \argmax_{\mathbf{u}_{j}\in\mathcal{P}_k} \frac{\mathbf{z}_i\cdot\mathbf{u}_{j}}{\|\mathbf{z}_i\|\|\mathbf{u}_{j}\|} \,,
\end{equation}
where $\cdot$ denotes the dot product. 
\begin{figure}[t]
    \centering
    \includegraphics[width=\linewidth]{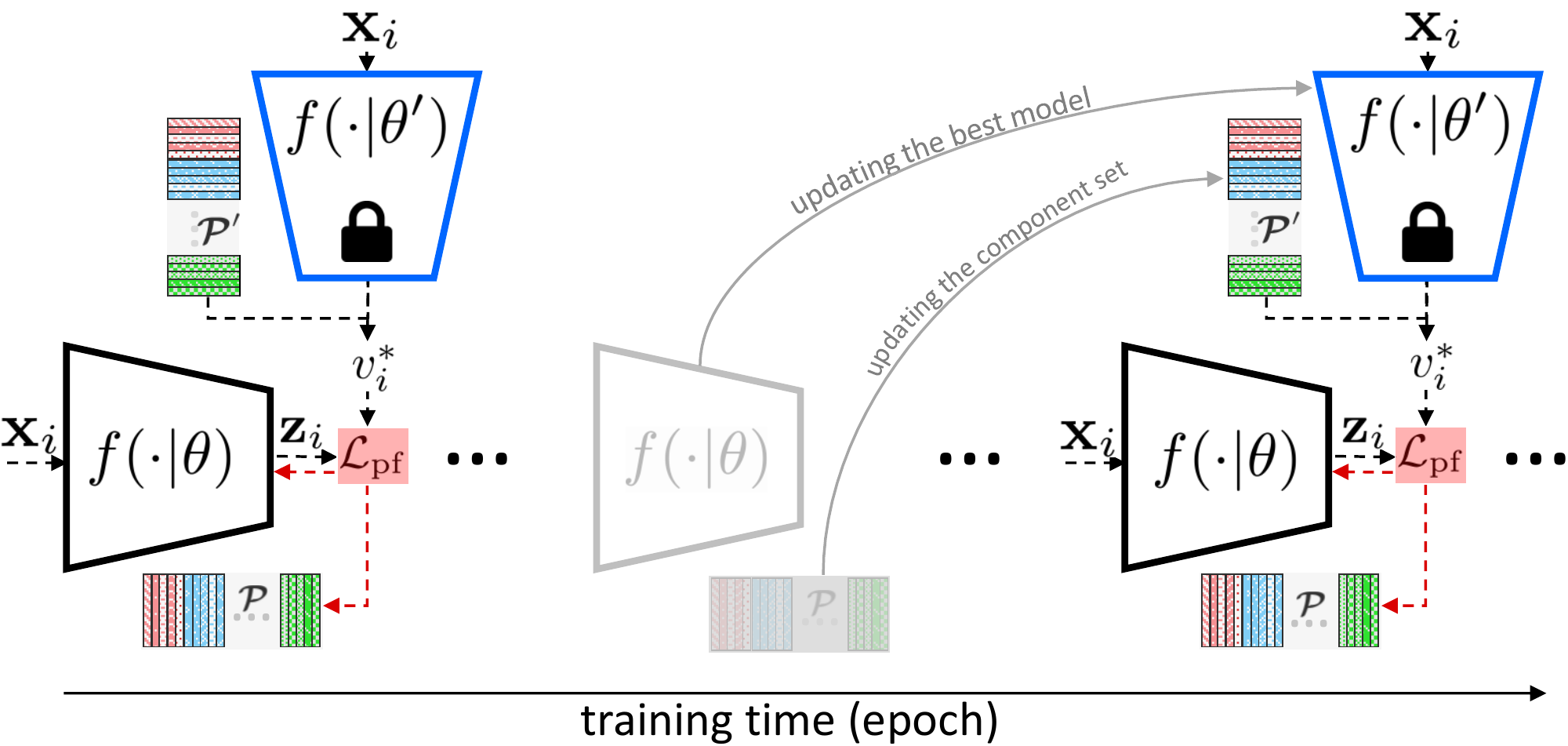}
    \caption{Progressive following training stage. $f(\cdot|\theta)$ is adapted using loss function $\mathcal{L}_\mathrm{pf}$ (eq.~\ref{eq:sp}) and supervised by a fixed copy of the best target model $f(\cdot|\theta')$ (in blue) and the corresponding mixture $\bm{\mathcal{P}}'$ after the initial training stage. The gradient (dashed red line) is backpropagated only through $f(\cdot|\theta)$ and $\bm{\mathcal{P}}$, while $f(\cdot|\theta')$ and $\bm{\mathcal{P}}'$ are kept fixed.  
    The target network and mixture $f(\cdot|\theta')$ and $\bm{\mathcal{P}}'$ are replaced by the best validated $f(\cdot|\theta)$ and $\mathcal{P}$ after $\alpha_3$ number of training steps with no improvement in validation.  
    The temperature factor $\tau$ (eq.~\ref{eq:Compscoreing}) decreases each time the target network is updated to create progressively more discriminative clusters.}
    \label{fig:progressive_steep_following}
\end{figure}
\begin{algorithm}[t]
    \SetAlgoLined
    {\small
        \KwData{pre-trained $f(\cdot|{\theta})$, pre-trained $\bm{\mathcal{P}}$, base set $\mathcal{X}^b$, validation set $\mathcal{X}^v$, patience $\alpha_2$, number of repetitions $\alpha_3$, temperature $\tau$, decreasing ratio $\gamma$, and error evaluation function $E(\cdot)$}
        \KwResult{Refined model $f(\cdot|\theta^\textrm{best})$ and mixture $\bm{\mathcal{P}}^\textrm{best}$}
        $\theta'\leftarrow\theta$; $\bm{\mathcal{P}}'\leftarrow\bm{\mathcal{P}}$; $\theta^\textrm{best}\leftarrow\theta$; $\bm{\mathcal{P}}^\textrm{best}\leftarrow\bm{\mathcal{P}}$; $s\leftarrow 0$\\
        \For{$t=1,2,\ldots,\alpha_3$}
        {
            \While{$s<\alpha_2$}
            { 
                \For{$(\mathbf{x}_i, \mathbf{y}_i)\in\mathcal{X}^b$}
                {
                    Evaluate $\mathbf{z}_i\leftarrow f(\mathbf{x}_i|\theta')$ %\\
                    %Evaluate 
                    , and $\mathbf{u}_i^*{}'$ 
                    %of $\mathbf{z}_i$ 
                    by eq.~\ref{eq:argmax_sp}\\ 
                    Update weights $\theta$ and mixture $\bm{\mathcal{P}}$ by backward error propagation from $\mathcal{L}_\mathrm{pf}$ (eq.~\ref{eq:sp})\\ 
                }
                \eIf{$E(\theta,\bm{\mathcal{P}}|\mathcal{X}^v) < E(\theta^\mathrm{best},\bm{\mathcal{P}}^\mathrm{best}|\mathcal{X}^v)$}
                {
                    $\theta^\textrm{best}\leftarrow\theta$; $\bm{\mathcal{P}}^\textrm{best}\leftarrow\bm{\mathcal{P}}$; $s\leftarrow 0$
                    %\todo{this never exits the loop, is this right? }
                }
                {
                    $s\leftarrow s+1$
                }
            } 
            Update target $\theta'\leftarrow\theta^\textrm{best}$ and mixture $\bm{\mathcal{P}}'\leftarrow\bm{\mathcal{P}}^\textrm{best}$\\ 
            Decrease temperature $\tau$ of eq.~\ref{eq:Compscoreing} as $\tau\leftarrow\gamma\tau$\\
        }
    }
    \caption{Progressive following. %\todo{Not sure where, but define $E(\cdot)$ somewhere.}
    }
    \label{alg:steep_following}
\end{algorithm} 
Based on this, the ``assignment'' loss function $\mathcal{L}_\mathrm{a}$ updates both $f(\cdot|\theta)$ and $\bm{\mathcal{P}}$ such that $\mathbf{z}_i$ is assigned to its most similar component $\mathbf{u}^*_i$:
\begin{equation} 
  \label{eq:loss-assign}  
  \mathcal{L}_{\mathrm{a}} = -\frac{1}{N} \sum_{i=1}^N  \log  p_{\theta}(v^*_i| \mathbf{z}_i, \bm{\mathcal{P}}) \,, 
\end{equation} 
where $N$ is the batch size and $v^*_i$ is the one-hot pseudo-label corresponding to $\mathbf{u}_i^*$. The gradient of eq.~\ref{eq:loss-assign} is back-propagated to both $f(\cdot|\theta)$ and the learned components $\bm{\mathcal{P}}$. 

As verified later (sec.~\ref{sec:ablativeAnalysis}), training solely on the assignment loss $\mathcal{L}_{\mathrm{a}}$ generally results in a single component $\mathbf{u}_i \in \mathcal{P}_k$ to be assigned to all training instances for class $k$, thereby effectively degrading the learned mixtures to a single mode. 
We compensate for this by adding a second loss function to encourage a diversity of components to be selected by enforcing $f(\cdot|\theta)$ to push the $\mathbf{z}_i$ values towards the centroid of the components corresponding to their associated labels $y_i$. 
For the centroid $\bm{c}_k = ({1}/{|\mathcal{P}_k|}) \sum_{\mathbf{u}_{j}\in\mathcal{P}_k}\mathbf{u}_{j}$ for base class $k$, and the set $\bm{\mathcal{C}} = \{\bm{c}_k\}_{k=1}^{N^b}$ of the centroids for base classes, we define the \emph{diversity} loss as: 
\begin{equation} 
  \label{eq:loss-div}  
  \mathcal{L}_{\mathrm{d}} = -\frac{1}{N} \sum_{i=1}^N \log  p_{\theta}(y_i| \mathbf{z}_i, \mathrm{sg}[\bm{\mathcal{C}}]) \,,
\end{equation}   
where $\mathrm{sg}$ stands for stopgradient, which blocks backpropagation over the variables it protects. The $\mathrm{sg}$ operator in eq.~\ref{eq:loss-div}  prevents the collapsing of all components of the $k$-th class $\mathcal{P}_k$ into a single point.  %  
Overall, the loss in this initial stage is the combination of eqs~\ref{eq:loss-assign} and~\ref{eq:loss-div}:
\begin{align} 
\label{eq:loss-it}  
\mathcal{L}_\mathrm{it} =  \mathcal{L}_{\mathrm{a}} + \mathcal{L}_{\mathrm{d}} \,.
\end{align}

\subsection{Progressive following}
\label{sec:progressive-following}
 
After the initial training of the feature extractor $f(\cdot|\theta)$ and mixture $\bm{\mathcal{P}}$, an intense competition is likely to arise for the assignment of the nearest components to each instance $\mathbf{z}_i$.
To illustrate this, suppose $\mathbf{\dot{u}}$ is assigned to $\mathbf{z}$ at iteration $t$. At the following iteration $t+1$, the simultaneous weight update to both $f(\cdot|\theta)$ and $\bm{\mathcal{P}}$ could cause another ${\mathbf{\ddot{u}}}$, in the vicinity of $\mathbf{\dot{u}}$ and $\mathbf{z}$, to be assigned as the nearest component of $\mathbf{z}$. 
Given the margin-based softmax (eq.~\ref{eq:Compscoreing}), $\mathbf{z}$ is pulled toward $\mathbf{\dot{u}}$ and pushed away from $\mathbf{\ddot{u}}$ at iteration $t$, and contradictorily steered in the opposite direction at the following iteration. 
As a result, this ``pull-push'' behavior stalls the improvement of $f(\cdot|\theta)$, preventing it from making further progress. 

To tackle this problem, we propose a progressive following stage that aim to break the complex dynamic of simultaneously determining nearest components while training the representation $f(\cdot|\theta)$ and mixture $\bm{\mathcal{P}}$.
The approach is detailed in algorithm~\ref{alg:steep_following} and shown in fig.~\ref{fig:progressive_steep_following}. 
Using the ``prime'' notation ($\theta^\prime$ and $\bm{\mathcal{P}}^\prime$ to specify the best feature extractor parameters and mixture component so far, resp.), the approach starts by taking a copy of $f(\cdot|\theta')$ and $\bm{\mathcal{P}}^\prime$, and by using them to determine the nearest component of each training instance:
\begin{equation} \label{eq:argmax_sp}
    \mathbf{u}{_i^*}^\prime = \argmax_{\mathbf{u}_j'\in\mathcal{P}_k'} \frac{\mathbf{z}'_i\cdot\mathbf{u}_j'}{\|\mathbf{z}'_i\|\|\mathbf{u}_j'\|} \,,
\end{equation}
where $\mathbf{z}'_i = f(\mathbf{x}_i|\theta')$. 
Since determining the labels does not depend on the learned parameters $\theta$ anymore, consistency in the assignment of nearest components is preserved, and the ``push-pull'' problem mentioned above is eliminated.

Since label assignments are fixed, the diversity loss (eq.~\ref{eq:loss-div}) is not needed anymore. Therefore, we can reformulate the progressive assignment loss function as:
\begin{equation} 
\label{eq:sp}  
    \mathcal{L}_{\mathrm{pf}} = -\frac{1}{N} \sum_{i=1}^N  \log  p_{\theta}({v_i^*}'| \mathbf{z}_i, \bm{\mathcal{P}}) \,, 
\end{equation} 
where $N$ is the batch size and $v{_i^*}'$ the pseudo-label associated to the nearest component $\mathbf{u}{_i^*}'$ found by eq.~\ref{eq:argmax_sp}.

After $\alpha_2$ updates to the representation with no decrease of the validation set error (function $E(\cdot)$ in algorithms \ref{algo:initial_train} and \ref{alg:steep_following}), the best network $f(\cdot|\theta')$ and mixture $\bm{\mathcal{P}}'$ are then replaced with the new best ones found on validation set, the temperature $\tau$ is decreased by a factor $\gamma<1$ to push the $\mathbf{z}$ more steeply towards their closest mixture component, and the entire procedure is repeated as shown in algorithm~\ref{alg:steep_following}. After a maximum number of $\alpha_3$ iterations is reached, the global best possible model $\theta^\textrm{best}$ and mixture ${\mathcal{P}}^\textrm{best}$ are obtained. Components that have no base class samples associated (i.e. never selected by eq.~\ref{eq:argmax_sp}) are simply discarded. This effectively adapts the mixture models to each base class distribution. 

In summary, the progressive following aims at solving the discussed pull-push behavior observed (see sec.~\ref{sec:ablativeAnalysis}). This stage applies a similar approach than in initial stage, with two significant differences: 1) the diversity loss $\mathcal{L}_\mathrm{d}$ is removed; and 2) label assignments are provided by a copy of the best model so far $f(\cdot|\theta^\prime)$ to stabilize the training.

\begin{table}[t]
\renewcommand{\tabcolsep}{2pt}
\centering 
\caption{Evaluation on miniImageNet in 5-way. Bold/blue is best/second, and $\pm$ is the 95\% confidence intervals in 600 episodes.} 
\begin{tabular}{ llccccccc}
    \bottomrule  
        & \textbf{Method}   
        & \textbf{\small Backbone} 
        & \textbf{1-shot}  
        & \textbf{5-shot}    
        \\ 
       \midrule
                     
            & ProtoNet~\cite{snell2017prototypical}   
            & {\small Conv4}
            & 49.42\scriptsize{ $\pm$ 0.78}      & 68.20\scriptsize{ $\pm$ 0.66}  
            \\
                     
            & MAML~\cite{finn2018probabilistic}       
            & {\small Conv4}
            & 48.07\scriptsize{ $\pm$ 1.75}      & 63.15\scriptsize{ $\pm$ 0.91}  
            \\
                     
            & RelationNet~\cite{sung2018learning}    
            & {\small Conv4}
            & 50.44\scriptsize{ $\pm$ 0.82}      & 65.32\scriptsize{ $\pm$ 0.70}    
            \\  
                    
            & Baseline++~\cite{chen2019closer}    
            & {\small Conv4}
            & 48.24\scriptsize{ $\pm$ 0.75} & 66.43\scriptsize{ $\pm$ 0.63}      
            \\
                  
            & IMP~\cite{allen2019infinite}
            & {\small Conv4}
            & 49.60\scriptsize{ $\pm$ 0.80}    & 68.10\scriptsize{ $\pm$ 0.80}  
            \\ 
            & MemoryNetwork~\cite{Cai_2018_CVPR}
            & {\small Conv4}
            & \textbf{53.37}\scriptsize{ $\pm$ 0.48}    & 66.97\scriptsize{ $\pm$ 0.35}  
            \\ 
            & Arcmax~\cite{Afrasiyabi_2020_ECCV}       
                & \small{Conv4} 
                & 51.90\scriptsize{ $\pm$0.79}      & 69.07\scriptsize{ $\pm$ 0.59} \\
            & Neg-Margin~\cite{Bin_2020_ECCV_margin_matter}       
                & \small{Conv4} 
                & {\color{blue}52.84}\scriptsize{ $\pm$0.76}      & {\color{blue}70.41}\scriptsize{ $\pm$0.66} 
           
           \\*[0.5em]  
            & MixtFSL (ours)
                & \small{Conv4} 
                & 52.82\scriptsize{ $\pm$0.63}      & \textbf{70.67}\scriptsize{ $\pm$0.57} 
            \\ 
            
        \midrule 
        %\multirow{5}{*}{\rotatebox{90}{meta learning}}    
                & DNS~\cite{Simon_2020_CVPR}  
                    & {\small RN-12}
                    & 62.64\scriptsize{ $\pm$0.66}           & 78.83\scriptsize{ $\pm$0.45}  
                    \\ 
                    
                & Var.FSL~\cite{zhang2019variational}      
                    & {\small RN-12}
                    & 61.23\scriptsize{ $\pm$0.26}           & 77.69\scriptsize{ $\pm$0.17}  
                    \\ 
                    
                 & MTL~\cite{sun2019meta}       
                    & {\small RN-12}
                    & 61.20\scriptsize{ $\pm$1.80}           & 75.50\scriptsize{ $\pm$0.80}  
                    \\
                    
                & SNAIL~\cite{mishra2017simple}     
                    & {\small RN-12}
                    & 55.71\scriptsize{ $\pm$0.99}            & 68.88\scriptsize{ $\pm$0.92}  
                    \\ 
                & AdaResNet~\cite{munkhdalai2018rapid}  
                    & {\small RN-12}
                    & 56.88\scriptsize{ $\pm$0.62}            & 71.94\scriptsize{ $\pm$0.57}  
                    \\
                    
                & TADAM~\cite{oreshkin2018tadam}     
                    & {\small RN-12}
                    & 58.50\scriptsize{ $\pm$0.30}            & 76.70\scriptsize{ $\pm$0.30}  
                    \\
                    
                & MetaOptNet~\cite{lee2019meta}      
                    & {\small RN-12}
                    & 62.64\scriptsize{ $\pm$0.61}               & 78.63\scriptsize{ $\pm$0.46} 
                    \\
                     
                & Simple~\cite{Tian_2020_ECCV_good}  %\todo{ECCV2020read this}   
                    & {\small RN-12}
                    & 62.02\scriptsize{ $\pm$0.63}           & 79.64\scriptsize{ $\pm$0.44}  
                    \\ 
                    
                & TapNet~\cite{yoon2019tapnet}    
                    & {\small RN-12}
                    & 61.65\scriptsize{ $\pm$0.15}            & 76.36\scriptsize{ $\pm$0.10}  
                  \\
                & Neg-Margin~\cite{Bin_2020_ECCV_margin_matter}      
                & \small{RN-12} 
                & {{\color{blue}63.85}}\scriptsize{ $\pm$0.76}      & {\color{blue}81.57}\scriptsize{ $\pm$0.56}      
                 \\*[0.5em] 
                  
         %----------------------------------MixtFSL(RN12)------------------------------------% 
            & MixtFSL (ours)   
                & \small{RN-12} 
                & \textbf{63.98}\scriptsize{ $\pm$0.79}       & \textbf{82.04}\scriptsize{ $\pm$0.49} 
            \\        
            \midrule            
                & MAML$^{\ddag}$~\cite{finn2017model}    
                    & {\small RN-18}  
                    & 49.61\scriptsize{ $\pm$0.92}           &  65.72\scriptsize{ $\pm$0.77}  
                    \\
                    
                & RelationNet$^{\ddag}$~\cite{sung2018learning}    
                    & {\small RN-18}
                    & 52.48\scriptsize{ $\pm$0.86}           &  69.83\scriptsize{ $\pm$0.68}  
                    \\
                    
                & MatchingNet$^{\ddag}$~\cite{vinyals2016matching}   
                    & {\small RN-18}
                    & 52.91\scriptsize{ $\pm$0.88}           &  68.88\scriptsize{ $\pm$0.69}  
                    \\
                    
                & ProtoNet$^{\ddag}$~\cite{snell2017prototypical}    
                    & {\small RN-18}
                    & 54.16\scriptsize{ $\pm$0.82}            & 73.68\scriptsize{ $\pm$0.65}  
                    \\

                & Arcmax~\cite{Afrasiyabi_2020_ECCV}       
                        & \small{RN-18} 
                        & 58.70\scriptsize{ $\pm$0.82}      & 77.72\scriptsize{ $\pm$0.51}   
                    \\
                    
                & Neg-Margin~\cite{Bin_2020_ECCV_margin_matter}    
                    & \small{RN-18} 
                    & {\color{blue}59.02}\scriptsize{ $\pm$0.81}      & \textbf{78.80}\scriptsize{ $\pm$0.54}       
                    \\*[0.5em]
                %----------------------------------MixtFSL(RN18)------------------------------------%
                    & MixtFSL (ours)
                        & \small{RN-18} 
                        & \textbf{60.11}\scriptsize{ $\pm$0.73}      & {\color{blue}77.76}\scriptsize{ $\pm$0.58}
                        \\  
               
        \midrule              
        & Act. to Param.~\cite{Qiao_2018_CVPR}  %{\color{red}RN-50}  
                    & {\small RN-50}
                    & 59.60\scriptsize{ $\pm$0.41}            & 73.74\scriptsize{ $\pm$0.19}  
                    \\
                    
        & SIB-inductive$^\S$\cite{hu2020empirical} 
                & {\small WRN}
                & 60.12\scriptsize{ \ \ \  \quad \quad}             & 78.17\scriptsize{ \ \ \  \quad \quad}  
                     \\ 
                     
            & SIB+IFSL~\cite{tang2020long}
                & {\small WRN}
                & 63.14\scriptsize{ $\pm$3.02}           & 80.05\scriptsize{ $\pm$1.88}  
                \\

            & LEO~\cite{rusu2018meta} 
                & {\small WRN}
                & 61.76\scriptsize{ $\pm$0.08}           & 77.59\scriptsize{ $\pm$0.12}  
                    \\

            & wDAE~\cite{gidaris2019generating}  
                & {\small WRN}
                & 61.07\scriptsize{ $\pm$0.15}           & 76.75\scriptsize{ $\pm$0.11}  
                    \\
                    
            & CC+rot~\cite{gidaris2019boosting}  
                & {\small WRN}
                & 62.93\scriptsize{ $\pm$0.45}           & 79.87\scriptsize{ $\pm$0.33}  
                    \\
                
            & Robust dist++~\cite{dvornik2019diversity}  
                & {\small WRN}
                & {\color{blue}63.28}\scriptsize{ $\pm$0.62}          & 81.17\scriptsize{ $\pm$0.43}  
                    \\

            & Arcmax~\cite{Afrasiyabi_2020_ECCV}       
                    & \small{WRN} 
                    & 62.68\scriptsize{ $\pm$0.76}      & 80.54\scriptsize{ $\pm$0.50}   
                \\
            & Neg-Margin~\cite{Bin_2020_ECCV_margin_matter}       
                & \small{WRN} 
                & 61.72\scriptsize{ $\pm$0.90}      & \textbf{81.79}\scriptsize{ $\pm$0.49}    
            \\*[0.5em]  
            %----------------------------------MixtFSL(RN18)------------------------------------
                & MixtFSL (ours)        
                    & \small{WRN}  
                    &  \textbf{64.31}\scriptsize{ $\pm$0.79}     & {\color{blue}81.66}\scriptsize{ $\pm$0.60} 
                    \\  
            \bottomrule    
    \end{tabular}  \\
    {\footnotesize $^{\ddag}$taken from~\cite{chen2019closer} \quad $^\S$confidence interval not provided }
    \label{tab:miniImageNet} 
\end{table}

\section{Experimental validation}
\label{sec:experimental-validation}
The following section presents the experimental validations of our novel mixture-based feature space learning (MixtFSL). We begin by introducing the datasets, backbones and implementation details. We then present experiments on object recognition, fine-grained and cross-domain classification. Finally, an ablative analysis is presented to evaluate the impact of decisions made in the design of MixtFSL.

\subsection{Datasets and implementation details}
\label{sec:datasets-implementation} 

\myparagraph{Datasets}  
Object recognition is evaluated using the miniImageNet~\cite{vinyals2016matching} and tieredImageNet~\cite{ren2018meta}, which are subsets of the ILSVRC-12 dataset~\cite{russakovsky2015imagenet}. miniImageNet contains 64/16/20 base/validation/novel classes respectively with 600 examples per class, and tieredImageNet~\cite{ren2018meta} contains 351/97/160 base/validation/novel classes. For fine-grained classification, we employ CUB-200-2011 (CUB)~\cite{wah2011caltech} which contains 100/50/50 base/validation/novel classes. For cross-domain, we train on the base and validation classes of miniImageNet, and evaluate on the novel classes of CUB.

\myparagraph{Backbones and implementation details} 
We conduct experiments using four different backbones: 1) Conv4, 2) ResNet-18~\cite{he2016deep}, 3) ResNet-12~\cite{he2016deep}, and 4) 28-layer Wide-ResNet (``WRN'')~\cite{sergey2016wide}. We used Adam~\cite{oreshkin2018tadam} and SGD with a learning rate of $10^{-3}$ to train Conv4 and ResNets and WRN, respectively. In SGD case, we used Nesterov with an initial rate of 0.001, and the weight decay is fixed as 5e-4 and momentum as 0.9.   
In all cases,  batch size is fixed to 128.
The starting temperature variable $\tau$ and margin $m$ (eq.~\ref{eq:Compscoreing} in sec.~\ref{sec:MixtFSL}) were found using the validation set (see supp. material). Components in $\bm{\mathcal{P}}$ are initialized with Xavier uniform~\cite{glorot2010understanding} ($\text{gain}=1$), and their number $N^k=15$ (sec.~\ref{sec:problemDefinition}), except for tieredImageNet where $N^k = 5$ since there is a much larger number of bases classes (351). 
A temperature factor of $\gamma=0.8$ is used in the progressive following stage. The early stopping thresholds of algorithms~\ref{algo:initial_train} and~\ref{alg:steep_following} are set to $\alpha_0=400$, $\alpha_1=20$, $\alpha_2=15$ and $\alpha_3=3$.

\begin{table}[t]
\renewcommand{\tabcolsep}{2pt}
\centering
\caption{Evaluation on tieredImageNet and FC100 in 5-way classification. Bold/blue is best/second best, and $\pm$ indicates the 95\% confidence intervals over 600 episodes.} 
\begin{tabular}{llcccc}  
        \toprule  
        & \textbf{Method}   
        & \textbf{\small Backbone} 
        & \textbf{1-shot}  
        & \textbf{5-shot}    
        \\  
        \midrule   
        \multirow{10}{*}{\rotatebox{90}{tieredImageNet}}   
                & DNS~\cite{Simon_2020_CVPR}  
                    & \small{RN-12}
                    & 66.22\scriptsize{ $\pm$0.75}           & 82.79\scriptsize{ $\pm$0.48}   
                    \\ 
                      
                & MetaOptNet~\cite{lee2019meta}      
                    & \small{RN-12} 
                    & 65.99\scriptsize{ $\pm$0.72}               & 81.56\scriptsize{ $\pm$0.53} 
                    \\
                     
                & Simple~\cite{Tian_2020_ECCV_good}  %\todo{ECCV2020read this}   
                    & \small{RN-12}  
                    & {\color{blue} 69.74}\scriptsize{ $\pm$0.72}           & {\color{blue}84.41}\scriptsize{ $\pm$0.55} 
                    \\ 
                    
                & TapNet~\cite{yoon2019tapnet}    
                    & \small{RN-12} 
                    & 63.08\scriptsize{ $\pm$0.15}            & 80.26\scriptsize{ $\pm$0.12}  
                    \\ 
                     
                & Arcmax$^*$~\cite{Afrasiyabi_2020_ECCV}      
                    & \small{RN-12}  
                    & 68.02\scriptsize{ $\pm$0.61}             & 83.99\scriptsize{ $\pm$0.62}  
                    \\*[.5em]  
            %\multirow{2}{*}{\rotatebox{90}{  }}   
                    & MixtFSL (ours) 
                    & \small{RN-12}  
                    & \textbf{70.97}\scriptsize{ $\pm$1.03}       & \textbf{86.16}\scriptsize{ $\pm$0.67}      
                    \\  
                
            \cmidrule(lr){2-5}  
            & Arcmax~\cite{Afrasiyabi_2020_ECCV} 
                    & \small{RN-18} 
                    & {\color{blue}65.08}\scriptsize{ $\pm$0.19}            & {\color{blue}83.67}\scriptsize{ $\pm$0.51} 
                    \\
                    
                & ProtoNet~\cite{snell2017prototypical}    
                    & \small{RN-18} 
                    & 61.23\scriptsize{ $\pm$0.77}            & 80.00\scriptsize{ $\pm$0.55}   
                    \\*[.5em]
                       
            %\cmidrule(lr){2-5}          
            & MixtFSL (ours)   
                    & \small{RN-18} 
                    & \textbf{68.61}\scriptsize{ $\pm$0.91}      & \textbf{84.08}\scriptsize{ $\pm$0.55}     
                    \\  
                
            \midrule       
            \multirow{8}{*}{\rotatebox{90}{FC100}}   

                & TADAM~\cite{oreshkin2018tadam} 
                & \small{RN-12} 
                & 40.1\scriptsize{ $\pm$ 0.40}       & {\color{blue}56.1\scriptsize{ $\pm$ 0.40}}  
                \\
                
                & MetaOptNet~\cite{lee2019meta}    
                & \small{RN-12} 
                & 41.1\scriptsize{ $\pm$ 0.60}       & 55.5\scriptsize{ $\pm$ 0.60}  
                \\
                
                & ProtoNet$^\dag$~\cite{snell2017prototypical}           
                & \small{RN-12} 
                & 37.5\scriptsize{ $\pm$ 0.60}       & 52.5\scriptsize{ $\pm$ 0.60}  
                \\
                
                & MTL~\cite{sun2019meta}       
                & \small{RN-12} 
                & {\color{blue}43.6\scriptsize{ $\pm$ 1.80}}       & 55.4\scriptsize{ $\pm$ 0.90}
                \\*[0.5em]

            %\cmidrule(lr){2-5}          
            & MixtFSL (ours)   
                    & \small{RN-12} 
                    & \textbf{44.89}\scriptsize{ $\pm$0.63}      & \textbf{60.70}\scriptsize{ $\pm$0.67}     
                    \\  
                
            \cmidrule(lr){2-5}  
            & Arcmax~\cite{Afrasiyabi_2020_ECCV} 
                & \small{RN-18}
                & 40.84\scriptsize{ $\pm$ 0.71}              & 57.02\scriptsize{ $\pm$ 0.63}   
                    \\*[.5em]
                       
            %\cmidrule(lr){2-5}          
            & MixtFSL (ours)   
                    & \small{RN-18} 
                    & \textbf{41.50}\scriptsize{ $\pm$0.67}      & \textbf{58.39}\scriptsize{ $\pm$0.62} 
                    %\test with another temp      
                    \\   
            \bottomrule       
    \end{tabular}  \\
    {\footnotesize $^*$our implementation  \quad $^{\dag}$taken from \cite{lee2019meta}}
    \label{tab:tieredIN-FC100}
\end{table}    
  
\subsection{Mixture-based feature space evaluations}  
\label{objectRec}

We first evaluate our proposed MixtFSL model on all four datasets using a variety of backbones. 

\myparagraph{miniImageNet} Table~\ref{tab:miniImageNet} compares our MixtFSL with several recent method on miniImageNet, with four backbones.  
MixtFSL provides accuracy improvements in all but three cases. %, and for 1-shot using Conv4, where it is practically tied with the best one. 
In the most of these exceptions, the method with best accuracy is Neg-Margin~\cite{Bin_2020_ECCV_margin_matter}, which is explored in more details in sec.~\ref{sec:extensions_neg_margin}.  
Of note, MixtFSL outperforms IMP~\cite{allen2019infinite} (sec.~\ref{sec:introduction} and \ref{sec:related-work}) by 3.22\% and 2.57\% on 1- and 5-shot respectively, thereby validating the impact of jointly learning the feature representation together with the mixture model.

\myparagraph{tieredImageNet and FC100} Table~\ref{tab:tieredIN-FC100} presents similar comparisons, this time on tieredImageNet and FC100. 
On both datasets and in both 1- and 5-shot scenarios, our method yields state-of-the-art results. In particular, MixtFSL results in classification gains of 3.53\% over Arcmax~\cite{Afrasiyabi_2020_ECCV} in 1-shot using RN-18, and 1.75\% over Simple~\cite{Tian_2020_ECCV_good} in 5-shot using ResNet-12 for tieredImageNet, and 1.29\% and 4.60\% over MTL~\cite{sun2019meta} for FC100 in 1- and 5-shot, respectively. 

\myparagraph{CUB} Table~\ref{tab:cub} evaluates our approach on CUB, both for fine-grained classification in 1- and 5-shot, and in cross-domain from miniImageNet to CUB for 5-shot using the ResNet-18. Here, previous work \cite{Bin_2020_ECCV_margin_matter} outperforms MixtFSL in the 5-shot scenario.  
We hypothesize this is due to the fact that either CUB classes are more unimodal than miniImageNet or that less examples per-class are in the dataset, which could be mitigated with self-supervised methods.

\begin{table}[t]
\renewcommand{\tabcolsep}{2pt}
\centering
\caption{
Fine-grained and on cross-domain from miniImageNet to CUB evaluation in 5-way using ResNet-18. Bold/blue is best/second, and $\pm$ is the 95\% confidence intervals on 600 episodes.} 
\begin{tabular}{rlcccc}  
        \toprule   
        &  & \multicolumn {2}{c}{\textbf{CUB}}    &\textbf{miniIN$\xrightarrow{}$CUB} \\
        & \textbf{Method} &\textbf{1-shot} &\textbf{5-shot}  &\textbf{5-shot}  \\ 
        \midrule          
 
            & GNN-LFT$^\diamond$~\cite{tseng2020cross}        %{\color{red}RN-10}
                %&{\footnotesize RN-10}   
                 & 51.51\scriptsize{ $\pm$0.8}     & 73.11\scriptsize{ $\pm$0.7}     
                     &--  
                     \\
        
                & Robust-20~\cite{dvornik2019diversity}   
                %& {\footnotesize RN-18}
                & 58.67\scriptsize{ $\pm$0.7}     & 75.62\scriptsize{ $\pm$0.5}  
                & --
                    \\      
                      
                & RelationNet$^\ddag$~\cite{sung2018learning}      
                    %& {\footnotesize RN-18} 
                    & 67.59\scriptsize{ $\pm$1.0}             & 82.75\scriptsize{ $\pm$0.6} 
                    & 57.71\scriptsize{ $\pm$0.7}
                     \\
                     
                & MAML$^\ddag$~\cite{finn2017model}    
                    %& {\footnotesize RN-18}
                    & 68.42\scriptsize{ $\pm$1.0}           & 83.47\scriptsize{ $\pm$0.6} 
                    & 51.34\scriptsize{ $\pm$0.7}
                    \\

                & ProtoNet$^{\ddag}$~\cite{snell2017prototypical}    
                    %& {\footnotesize RN-18}
                    & 71.88\scriptsize{ $\pm$0.9}             & {\color{blue}86.64}\scriptsize{ $\pm$0.5} 
                    & 62.02\scriptsize{ $\pm$0.7}
                    \\
                    
                & Baseline++~\cite{chen2019closer} 
                    %& {\footnotesize RN-18}
                    & 67.02\scriptsize{ $\pm$0.9}      &  83.58\scriptsize{ $\pm$0.5}
                    & 64.38\scriptsize{ $\pm$0.9}
                    \\ 
                
                & Arcmax~\cite{Afrasiyabi_2020_ECCV} 
                    %& {\footnotesize RN-18}
                    & 71.37\scriptsize{ $\pm$0.9}      & 85.74\scriptsize{ $\pm$0.5}
                    & 64.93\scriptsize{ $\pm$1.0}
                    \\
                    
                & Neg-Margin~\cite{Bin_2020_ECCV_margin_matter}  %\todo{ECCV2020: read this}   
                    %& {\footnotesize RN-18} 
                    & {\color{blue}72.66}\scriptsize{ $\pm$0.9}           & \textbf{89.40}\scriptsize{ $\pm$0.4}  
                    & {\color{blue}67.03}\scriptsize{ $\pm$0.8}
                    \\*[.5em] 
                     
            %\cmidrule(lr){2-5}           
                & MixtFSL (ours)  
                    %& {\footnotesize RN-18}
                    & \textbf{73.94}\scriptsize{ $\pm$1.1}      & 86.01\scriptsize{ $\pm$0.5} 
                    & \textbf{68.77}\scriptsize{ $\pm$0.9}
                    \\  
        \bottomrule       
    \end{tabular}  \\
    {\footnotesize $^\ddag$taken from \cite{tang2020long} \quad   $^\diamond$backbone is ResNet-10}
    \label{tab:cub}
\end{table}

\begin{figure} 
    \centering
    \footnotesize
    \setlength{\tabcolsep}{1pt}
    \begin{tabular}{cccc} 
    \includegraphics[width=0.3\linewidth, angle=90]{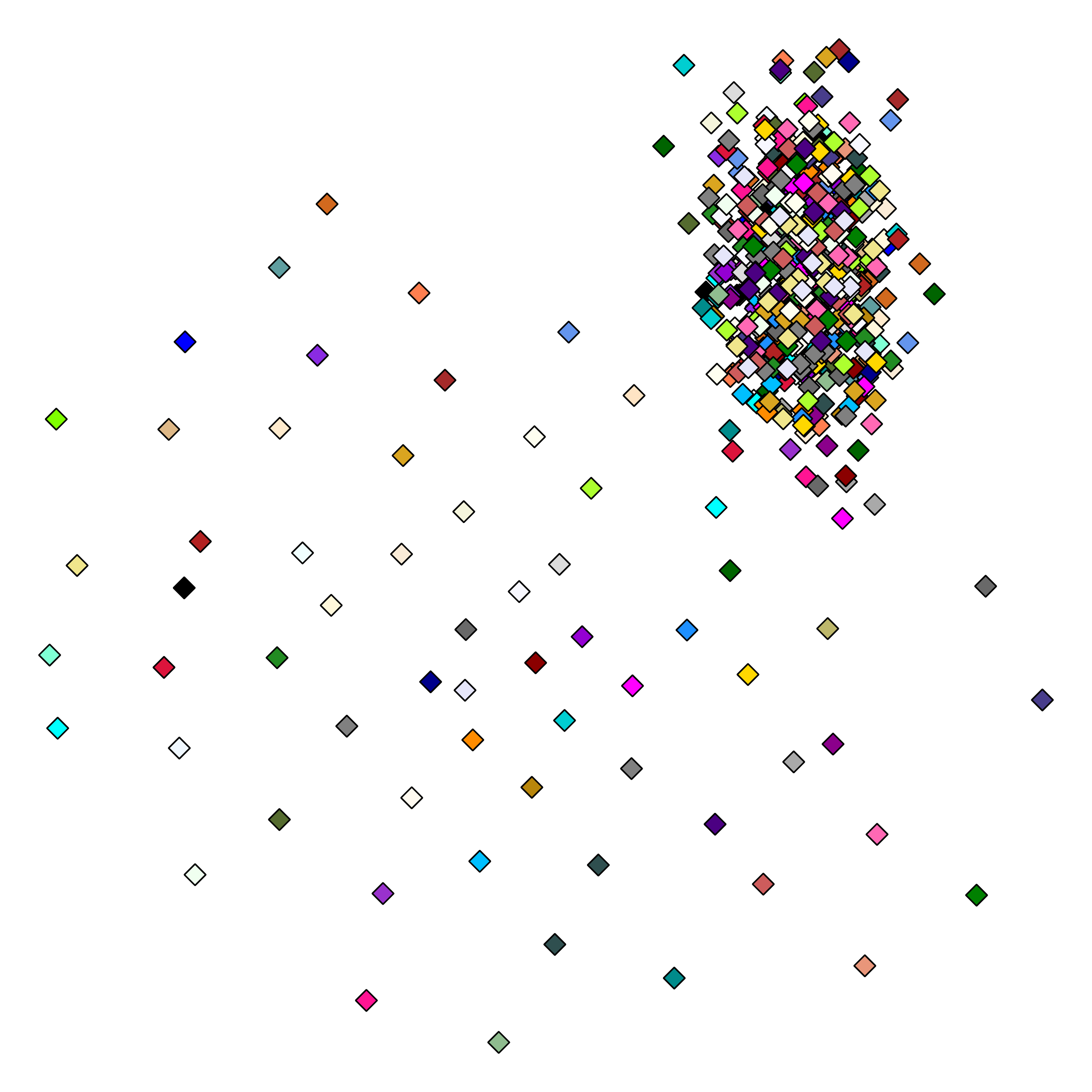} & 
    \includegraphics[width=0.3\linewidth, angle=90]{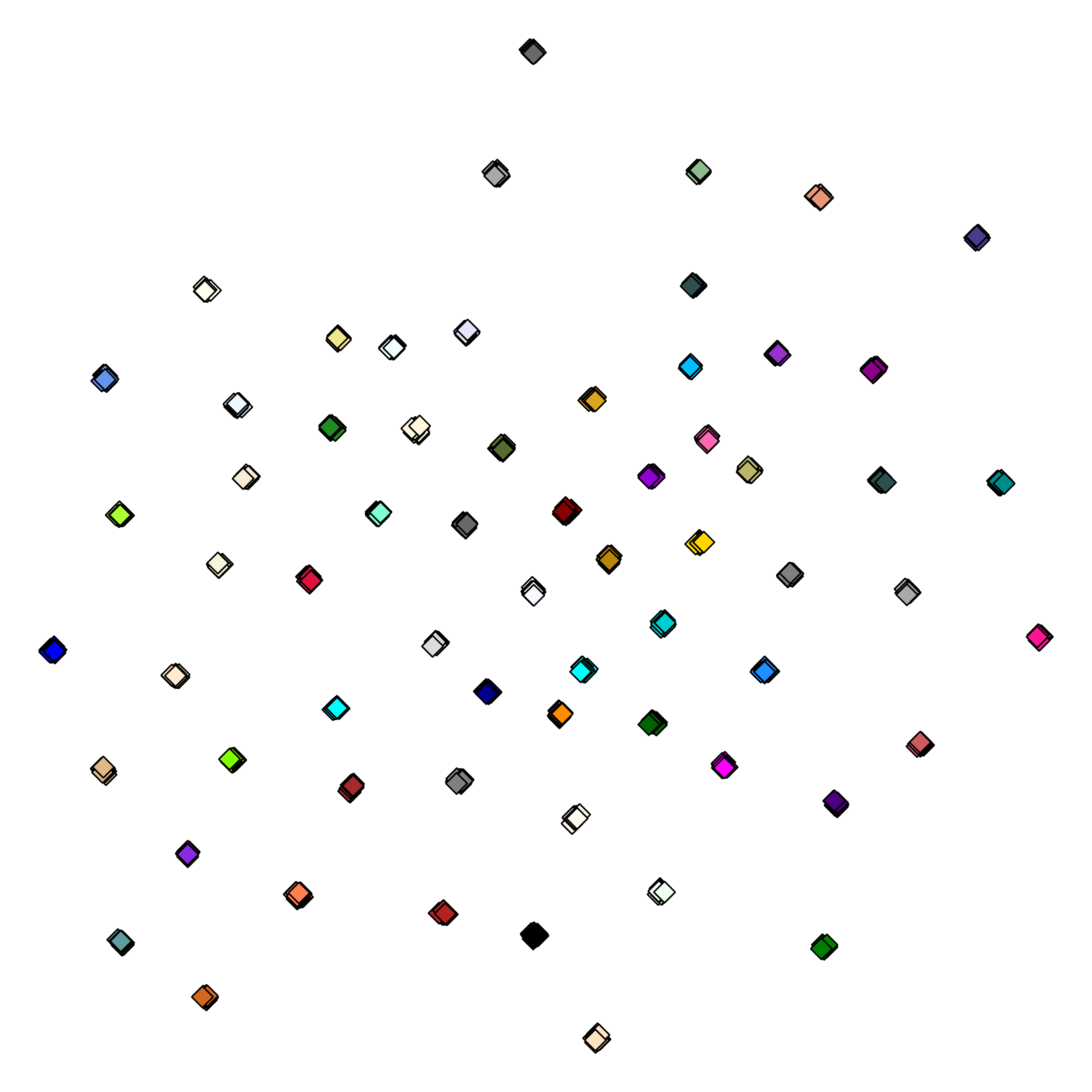} & 
    \includegraphics[width=0.3\linewidth, angle=0]{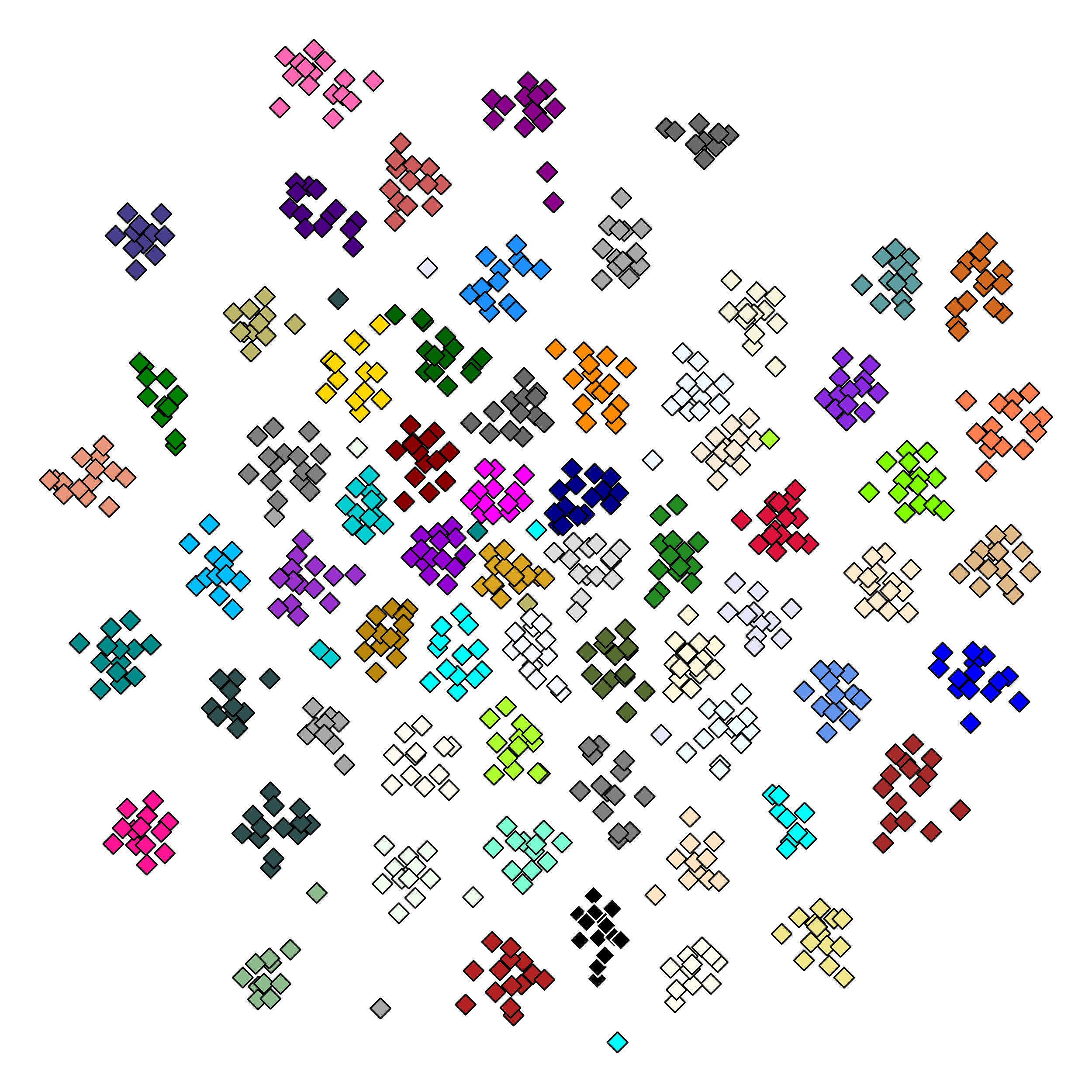} &   \\ 
    (a) without $\mathcal{L}_d$
    & (b) $\mathcal{L}_d$ without $\mathrm{sg}$
    & (c) $\mathcal{L}_d$ with $\mathrm{sg}$ \\       
    \end{tabular}
    \caption{t-SNE of mixture components (RN-12, miniImageNet). } 
    \label{fig:tsne_ld}
\end{figure}

\subsection{Ablative analysis}
\label{sec:ablativeAnalysis}

Here, we perform ablative experiments to evaluate the impact of two design decisions in our approach. 

\begin{figure} 
    \centering
    \footnotesize
    \setlength{\tabcolsep}{1pt}
    \begin{tabular}{ccc} 
    \includegraphics[width=0.4\linewidth]{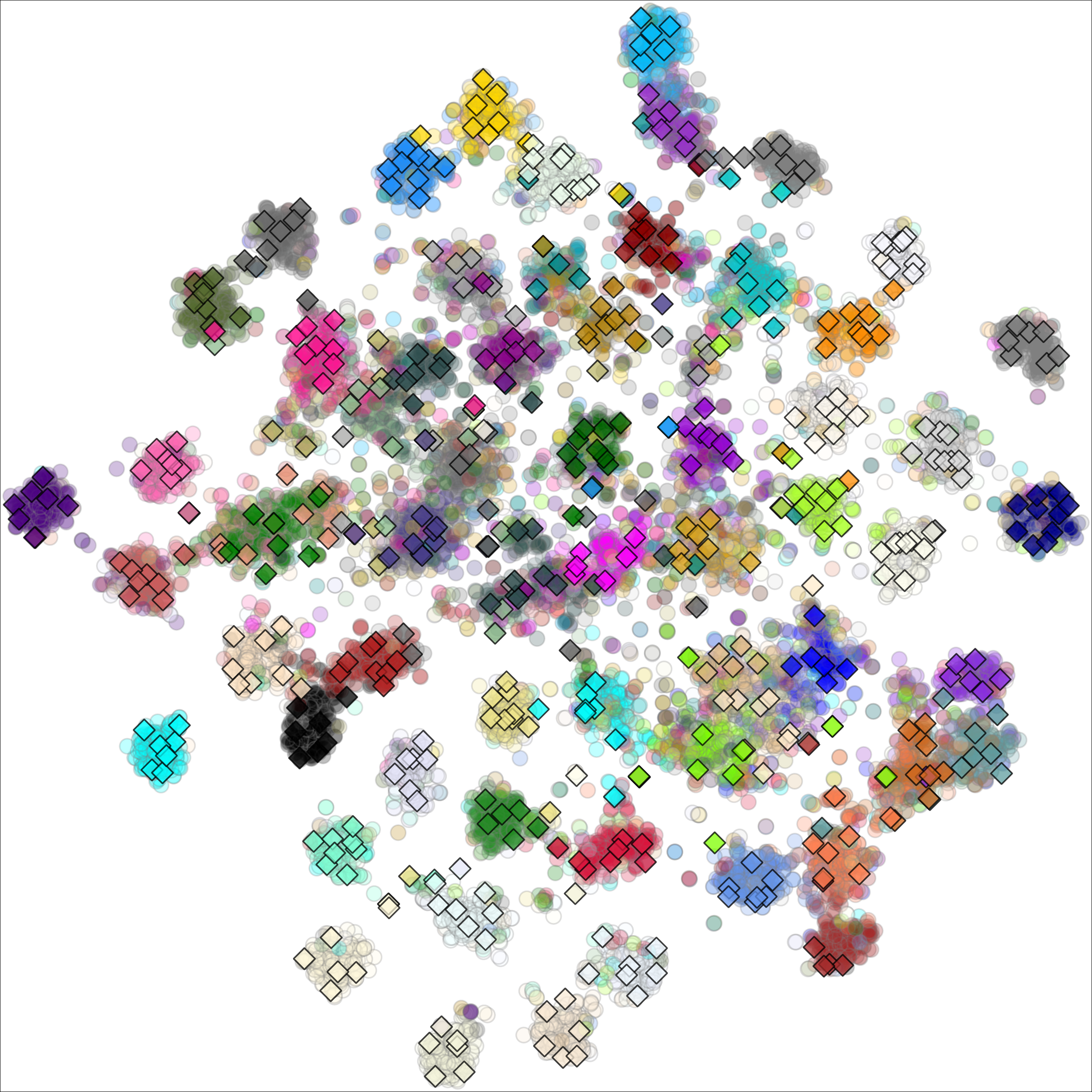} & 
    \includegraphics[width=0.4\linewidth]{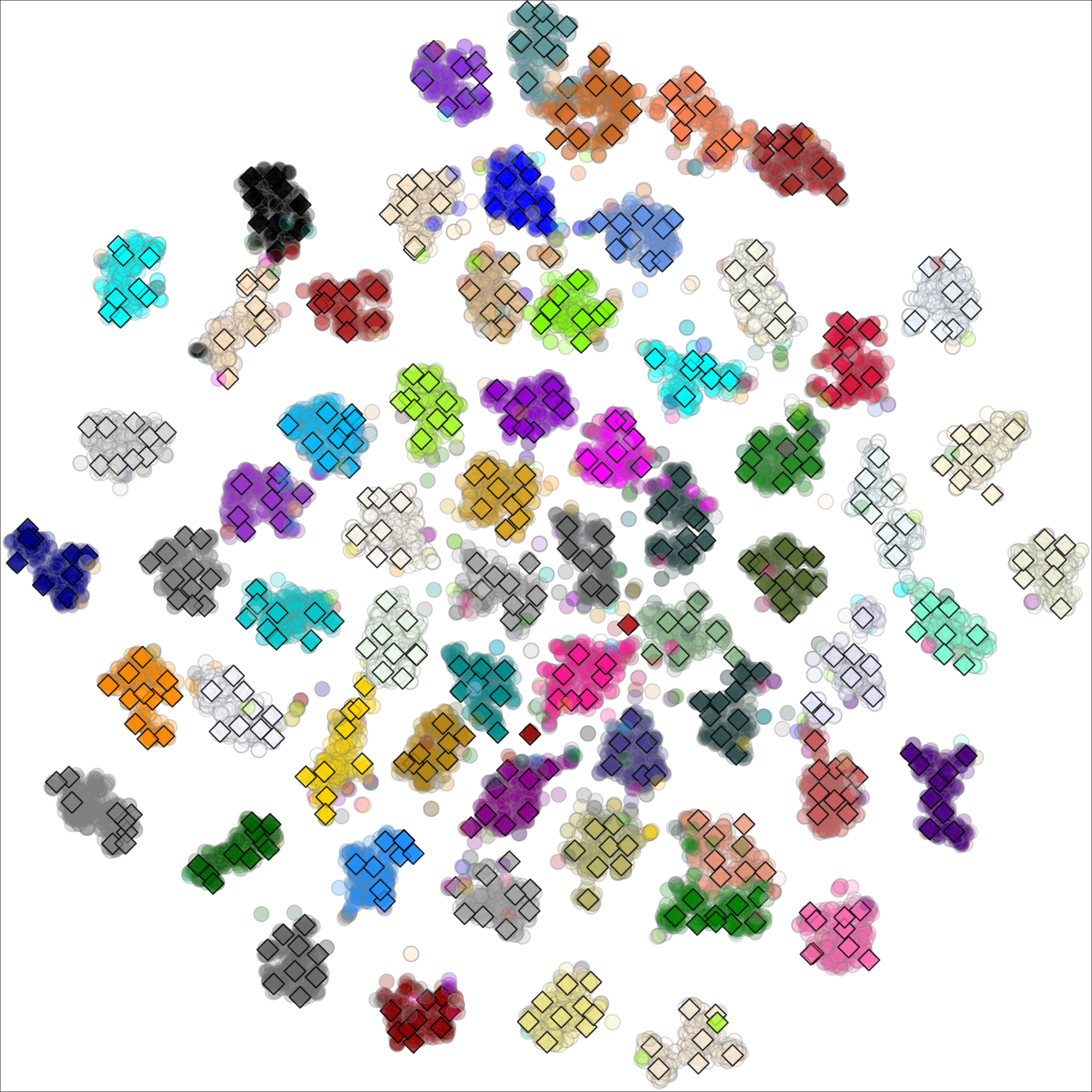} &   \\ 
    (a) after initial training  & 
    (b) after progressive following  \\       
    \end{tabular}
    \vspace{0.5em}
    \caption{t-SNE~\cite{maaten2008visualizing} visualization of the learned feature embedding (circles) and mixture components (diamonds), after the (a) initial training and (b) progressive following stages. Results are obtained with the ResNet-12 and points are color-coded by base class. }
    \label{fig:tsne_InitialVsProgressive}
\end{figure}

\myparagraph{Initial training vs progressive following} 
% We present in table~\ref{tab:modules} a quantitative ablation study
%\footnote{Numbers here are lower than in the paper since they are computed on the validation rather than the test classes and on 150 instead of 400 epochs.}.  
Fig.~\ref{fig:tsne_ld} illustrates the impact of loss functions qualitatively. 
% These will replace fig.~4 in the final paper. % (which will go to supp. mat.). 
Using only $\mathcal{L}_\mathrm{a}$ causes a single component to dominate while the others are pushed far away (big clump in fig.~\ref{fig:tsne_ld}a) and is equivalent to the baseline (table~\ref{tab:modules}, rows 1--2). Adding $\mathcal{L}_\mathrm{d}$ \emph{without} the $\mathrm{sg}$ operator minimizes the distance between the $\mathbf{z}_i$'s to the centroids, resulting in the collapse of all components in $\mathcal{P}_k$ into a single point (fig.~\ref{fig:tsne_ld}b). $\mathrm{sg}$ prevents the components (through their centroids) from being updated (fig.~\ref{fig:tsne_ld}c), which results in improved performance in the novel domain (t.~\ref{tab:modules}, row 3). Finally, $\mathcal{L}_\mathrm{pf}$ further improves performance while bringing stability to the training (t.~\ref{tab:modules}, row 4). 
Beside,
Fig.~\ref{fig:tsne_InitialVsProgressive} presents a t-SNE~\cite{maaten2008visualizing} visualization of base examples and their associated mixture components. Compared to initial training, the network at the end of progressive following stage results in an informative feature space with the separated base classes.

\begin{figure}
    \centering
    \footnotesize
    \setlength{\tabcolsep}{1pt}
    \begin{tabular}{cc} 
    \includegraphics[width=0.46\linewidth]{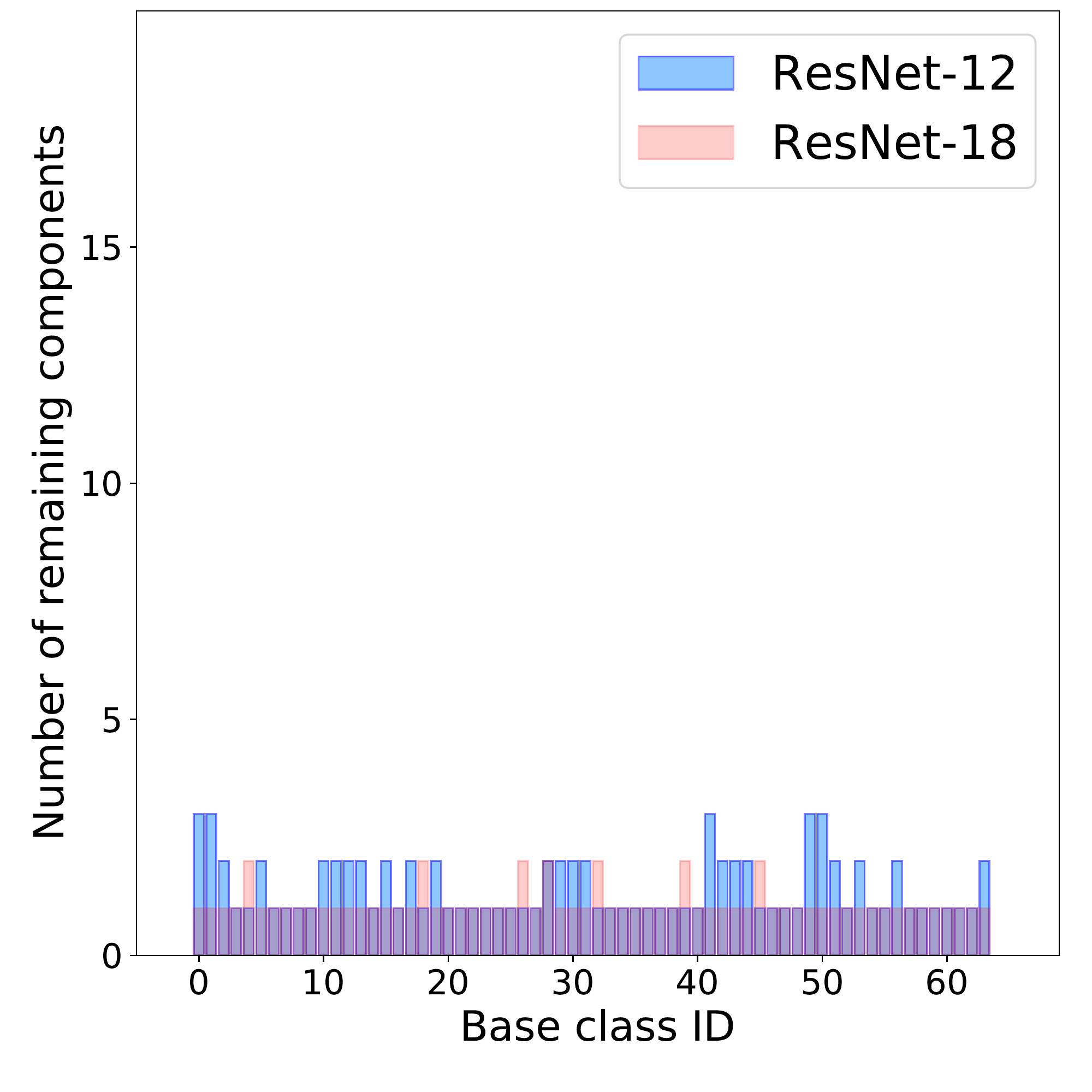} &
    \includegraphics[width=0.46\linewidth]{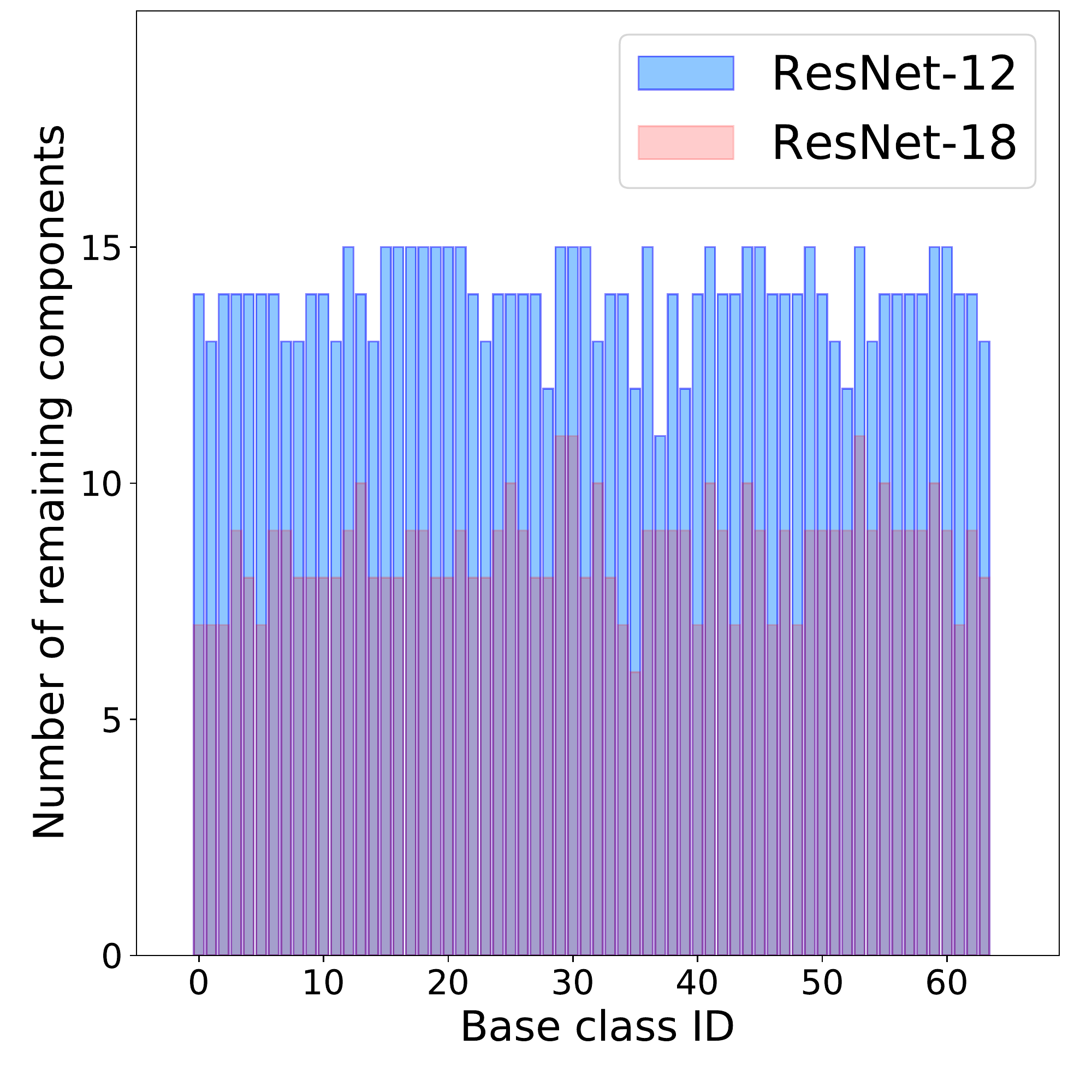} \\
    (a) without $\mathcal{L}_\mathrm{d}$ &
    (b) with $\mathcal{L}_\mathrm{d}$ \\
    \end{tabular}
    \caption{Number of remaining components after training for each of the miniImageNet base classes (a) without and (b) with the diversity loss $\mathcal{L}_\mathrm{d}$ (eq.~\ref{eq:loss-div}) using ResNet-12 and ResNet-18.
    The loss is critical to model the multimodality of base classes. 
    }
    \label{fig:histogram}
\end{figure}

\myparagraph{Diversity loss $\mathcal{L}_\mathrm{d}$} Fig.~\ref{fig:histogram} presents the impact of our diversity loss $\mathcal{L}_\mathrm{d}$ (eq.~\ref{eq:loss-div}) by showing the number of remaining components after optimization (recall from sec.~\ref{sec:progressive-following} that components assigned to no base sample are discarded after training). Without $\mathcal{L}_\mathrm{d}$ (fig.~\ref{fig:histogram}a), most classes are represented by a single component. Activating $\mathcal{L}_\mathrm{d}$ results in a large number of components having non-zero base samples, thereby results in the desired mixture modeling (fig.~\ref{fig:histogram}b).

\begin{table}[tb] 
\footnotesize
\centering
\caption[]{Validation set accuracy of miniImageNet on 150 epochs.  } 
\begin{tabular}{lcccc}         %& & \multicolumn{2}{c}{ResNet-18}   \\  %\cline{3-4}  \cline{6-7}    
% adversarial
    \toprule
           & \multicolumn{2}{c}{\textbf{RN-12}} & \multicolumn{2}{c}{\textbf{RN-18}} \\
    \textbf{Method} & \textbf{1-shot}  & \textbf{5-shot} & \textbf{1-shot} & \textbf{5-shot} \\ 
    \midrule
    Baseline                                                      &56.55       &72.68     &55.38       &72.81 \\
    Only $\mathcal{L}_\mathrm{a}$                                 &56.52       &72.78     &55.55       &72.67 \\
    Init. tr. ($\mathcal{L}_\mathrm{a} + \mathcal{L}_\mathrm{d}$) &57.88       &73.94     &56.18       &69.43\\
    Prog. fol. ($\mathcal{L}_\mathrm{a} + \mathcal{L}_\mathrm{d} + \mathcal{L}_\mathrm{pf}$)                 &58.60       &76.09  &57.91       &73.00\\
    \bottomrule
\end{tabular}
\label{tab:modules}
\vspace{-1em}
\end{table}

\myparagraph{Margin in eq.~\ref{eq:Compscoreing}}
\label{sec:extensions_neg_margin}
As in \cite{Afrasiyabi_2020_ECCV} and \cite{Bin_2020_ECCV_margin_matter}, our loss function (eq.~\ref{eq:Compscoreing}) uses a margin-based softmax function modulated by a temperature variable $\tau$. In particular, \cite{Bin_2020_ECCV_margin_matter} suggested that a negative margin $m < 0$ improves accuracy. Here, we evaluate the impact of the margin $m$, and demonstrate in table~\ref{tab:margin-ablation-shoter} that MixtFSL does not appear to be significantly affected by its sign.   

\begin{table}[t]
\renewcommand{\tabcolsep}{2pt}
\centering
\caption{Margin ablation using miniImageNet in 5-way classification. Bold/blue is best/second best, and $\pm$ indicates the 95\% confidence intervals over 600 episodes.} 

\begin{tabular}{rlccc}  
        \toprule  
        & \textbf{Method}   
        & \textbf{\small Backbone} 
        & \textbf{1-shot}  
        & \textbf{5-shot}    
        \\  
        \midrule
         %----------------------------------MixtFSL(RN12)------------------------------------% 
            & MixtFSL-Neg-Margin    
                & \small{RN-12} 
                & \textbf{63.98}\scriptsize{ $\pm$0.79}       & \textbf{82.04}\scriptsize{ $\pm$0.49} 
            \\
            &  MixtFSL-Pos-Margin    
                & \small{RN-12} 
                & 63.57\scriptsize{ $\pm$0.00}      & 81.70\scriptsize{ $\pm$0.49} 
            \\%*[0.5em]   
            
        \midrule      
            & MixtFSL-Neg-Margin
                & \small{RN-18} 
                & \textbf{60.11}\scriptsize{ $\pm$0.73}      & {\color{blue}{77.76}}\scriptsize{ $\pm$0.58}
                \\  
            & MixtFSL-Pos-Margin        
                & \small{RN-18} 
                & 59.71\scriptsize{ $\pm$0.76}     & {77.59}\scriptsize{ $\pm$0.58} 
                \\%*[0.5em]    
                 
         \bottomrule 
        
    \end{tabular}  \\  
    \label{tab:margin-ablation-shoter} 
\end{table}

\section{Extensions}
\label{sec:extensions}

We present extensions of our approach that make use of two recent works:   
the associative alignment of Afrasiyabi~\etal~\cite{Afrasiyabi_2020_ECCV}, and Ordinary Differential Equation (ODE) of Xu~\etal~\cite{xu2021learning}. In both cases, employing their strategies within our framework yields further improvements, demonstrating the flexibility of our MixtFSL.%representation learning approach.

% We report results on standard backbones with \emph{no additional architectural changes} as opposed to these works:~\cite{li2021learning} adds stack of 3 conv. layers as a pre-backbone to train (SElayer, CSEI and TSFM) modules,~\cite{zhao2021looking} uses a pre-trained ResNet-12 to train a ``Cross Non-local Network'',~\cite{xu2021learning} modifies the ResNet-12 backbone and trains a Neural ODE,~\cite{fei2020melr} adds an attention module with 1.64M params to ResNet-12, and~\cite{zhang2020iept} leverages a large set of additional unlabeled data. 

\subsection{Associative alignment~\cite{Afrasiyabi_2020_ECCV}}
\label{sec:alignment}  

Two changes are necessary to adapt our  MixtFSL to exploit the ``centroid alignment'' of Afrasiyabi~\etal~\cite{Afrasiyabi_2020_ECCV}. 
First, we employ the learned mixture model $\bm{\mathcal{P}}$ to find the related base classes. This is both faster and more robust than \cite{Afrasiyabi_2020_ECCV} who rely on the base samples themselves. Second, they used a %single linear 
classification layer $\mathbf{W}$ in $c(\mathbf{x}|\mathbf{W})\equiv\mathbf{W}^\top f(\mathbf{x}|\theta)$ (followed by softmax). Here, we use two heads ($\mathbf{W}^b$ and $\mathbf{W}^n$), to handle base and novel classes separately.

\begin{table}[t]
\renewcommand{\tabcolsep}{2pt}
\centering 
\caption{Comparison of our MixtFSL with alignment (MixtFSL-Align) in 5-way classification. Here, bold is the best performance. %Here, bold is best and $\pm$ indicates the 95\% confidence intervals over 600 episodes.
} 
\begin{tabular}{llccccccc}
    \bottomrule  
        & \textbf{Method}   
        & \textbf{\small Backbone} 
        & \textbf{1-shot}  
        & \textbf{5-shot}    
        \\ 
       \midrule 
       \multirow{6}{*}{\rotatebox{90}{\hspace{2em}miniIN}}         
                    
                & Cent. Align.$^*$~\cite{Afrasiyabi_2020_ECCV}  
                    & {\small RN-12}
                    &{63.44}\scriptsize{ $\pm$0.67}          &{80.96}\scriptsize{ $\pm$0.61}  
                    \\
                     
                & MixtFSL-Align. (ours)
                    & \small{RN-12}  
                     & \textbf{64.38}\scriptsize{ $\pm$0.73}       & \textbf{82.45}\scriptsize{ $\pm$0.62} 
                    \\  
                     
            \cmidrule(lr){2-5}           
                    
            & Cent. Align.$^*$~\cite{Afrasiyabi_2020_ECCV}  
                    & {\small RN-18}
                    &59.85\scriptsize{ $\pm$0.67}          &80.62\scriptsize{ $\pm$0.72}  
                    \\
                    
            & MixtFSL-Align. (ours)
                    & \small{RN-18}  
                    & \textbf{60.44}\scriptsize{ $\pm$1.02}       & \textbf{81.76}\scriptsize{ $\pm$0.74} 
                    \\   
        \midrule    
        %\bottomrule 
        \multirow{6}{*}{\rotatebox{90}{\hspace{1.5em}tieredIN}}
                    
            & Cent. Align.$^*$~\cite{Afrasiyabi_2020_ECCV}  
                    & {\small RN-12}
                    &{71.08}\scriptsize{ $\pm$0.93}          &{86.32}\scriptsize{ $\pm$0.66}  
                    \\
             
            & MixtFSL-Align. (ours)
                    & \small{RN-12}  
                    & \textbf{71.83}\scriptsize{ $\pm$0.99}       & \textbf{88.20}\scriptsize{ $\pm$0.55}      
                    \\ 
                    
        \cmidrule(lr){2-5}                 
                    
            & Cent. Align.$^*$~\cite{Afrasiyabi_2020_ECCV}  
                    & {\small RN-18}
                    &69.18\scriptsize{ $\pm$0.86}          &\textbf{85.97}\scriptsize{ $\pm$0.51}  \\  
                    
             & MixtFSL-Align. (ours)
                    & \small{RN-18}  
                    & \textbf{69.82}\scriptsize{ $\pm$0.81}       & 85.57\scriptsize{ $\pm$0.60}   
                    \\  
                    
        \bottomrule      
    \end{tabular}  \\
    {\footnotesize $^*$ our implementation}
    %\cite{tang2020long} \quad   $^{\dag}$ our implementation \quad    
    \label{tab:alignment}
\end{table}

\myparagraph{Evaluation} We evaluate our adapted alignment algorithm on the miniImageNet and tieredImageNet using the RN-18 and RN-12. Table~\ref{tab:alignment} presents our MixtFSL and  MixtFSL-alignment (MixtFSL-Align.) compared to \cite{Afrasiyabi_2020_ECCV} for the 1- and 5-shot (5-way) classification problems. Employing  MixtFSL improves over the alignment method of~\cite{Afrasiyabi_2020_ECCV} in all cases except in 5-shot (RN-18) on tieredImageNet, which yields slightly worse results. However, our  MixtFSL results in gain up to 1.49\% on miniImageNet and 1.88\% on tieredImageNet (5-shot, RN-12). To ensure a fair comparison, we reimplemented the approach proposed in \cite{Afrasiyabi_2020_ECCV} using our framework.

\myparagraph{Forgetting} Aligning base and novel examples improves classification accuracy, but may come at the cost of forgetting the base classes. Here, we make a comparative evaluation of this ``remembering'' capacity between our approach and that of Afrasiyabi~\etal~\cite{Afrasiyabi_2020_ECCV}. To do so, we first reserve 25\% of the base examples from the dataset, and perform the entire training on the remaining 75\%. After alignment, we then go back to the reserved classes and evaluate whether the trained models can still classify them accurately. Table~\ref{tab:forgetEvaluation} presents the results on miniImageNet. It appears that Afrasiyabi~\etal~\cite{Afrasiyabi_2020_ECCV} suffers from catastrophic forgetting with a loss of performance ranging from 22.1--33.5\% in classification accuracy. Our approach, in contrast, effectively remembers the base classes with a loss of only 0.5\%, approximately.

% \subsection{Ordinary Differential Equation~\cite{xu2021learning}}
\subsection{Combination with recent and concurrent works} 
\label{sec:ODE} 

Several recent and concurrent works~\cite{li2021learning,zhao2021looking,xu2021learning,fei2020melr} present methods which achieves competitive---or even superior---performance to that of MixtFSL presented in table~\ref{tab:miniImageNet}. They achieve this through improvements in neural network architectures: \cite{li2021learning} adds a stack of 3 convolutional layers as a pre-backbone to train other modules (SElayer, CSEI and TSFM), \cite{zhao2021looking} uses a pre-trained RN-12 to train a ``Cross Non-local Network'', and \cite{fei2020melr} adds an attention module with 1.64M parameters to the RN-12 backbone. 
Xu~\etal~\cite{xu2021learning} also modify the RN-12 and train an adapted Neural Ordinary Differential Equation (ODE), which consists of a dynamic meta-filter and adaptive alignment modules. The aim of the extra alignment module in~\cite{xu2021learning} is to perform channel-wise adjustment besides the spatial-level adaptation. In contrast to these methods, we emphasize that as opposed to these works, all MixtFSL results presented throughout the paper have been obtained with standard backbones \emph{without additional architectural changes}.

Since this work focuses on representation learning, our approach is thus orthogonal---and can be combined---to other methods which contain additional modules. To support this point, table~\ref{tab:extra_modules} combines MixtFSL with the ODE approach of Xu~\etal~\cite{xu2021learning} (MixtFSL-ODE) and shows that the resulting combination results in a gain of 0.85\% and 1.48\% over~\cite{xu2021learning} in 1- and 5-shot respectively.

\begin{table}[t]
\renewcommand{\tabcolsep}{2pt}
\centering
\caption{Evaluation of the capacity to remember base classes before and after alignment. Evaluation performed on miniImageNet in 5-way image classification. Numbers in () indicate the change in absolute classification accuracy compared to before alignment.}
\begin{tabular}{lcll}  
        \toprule  
        \textbf{Method} &    
        \textbf{\small Backbone} & 
        \multicolumn{1}{c}{\textbf{1-shot}} & 
        \multicolumn{1}{c}{\textbf{5-shot}} \\
        \midrule
        \cite{Afrasiyabi_2020_ECCV} before align.   & {\small RN-12} & 96.17 & 97.49 \\
        \cite{Afrasiyabi_2020_ECCV} after align.    & {\small RN-12} & 65.47 (-30.7) & 75.37 (-22.12) \\*[0.5em]
        ours before align.      & {\small RN-12} & 96.83 & 98.06 \\
        ours after align.       & {\small RN-12} & 96.27 (-0.6) & 98.11 (+0.1) \\
        \midrule
        \cite{Afrasiyabi_2020_ECCV} before align. 	& {\small RN-18} & 91.56 & 90.72 \\
        \cite{Afrasiyabi_2020_ECCV} after align.  	& {\small RN-18} & 58.02 (-33.5) & 62.97 (-27.8)\\*[0.5em]
        ours before align.      & {\small RN-18} & 97.46 & 98.16 \\
        ours after align.       & {\small RN-18} & 97.20 (-0.3) & 97.65 (-0.5) \\
         \bottomrule  
    \end{tabular}  \\  
    \label{tab:forgetEvaluation} 
\end{table}

\begin{table}[tb] 
\centering
\caption[]{Combining MixtFSL with the ODE appraoch of Xu~\etal~\cite{xu2021learning} (MixtFSL-ODE) in 5-way on miniImageNet using RN-12.} 
\begin{tabular}{lcc}          
% adversarial
    \toprule
    \textbf{Method} & \textbf{1-shot}  & \textbf{5-shot}   \\ 
    \midrule
    ODE~\cite{xu2021learning}                                 &67.76\scriptsize{ $\pm$0.46}       &82.71\scriptsize{ $\pm$0.31}    \\
    MixtFSL-ODE                      &\textbf{68.61}\scriptsize{ $\pm$0.73}       &\textbf{84.19}\scriptsize{ $\pm$0.44}     \\     
    \bottomrule
\end{tabular}
\label{tab:extra_modules}
\end{table}

\section{Discussion}
\label{sec:discussion}
This paper presents the idea of Mixture-based Feature Space Learning (MixtFSL) for improved representation learning in few-shot image classification. It proposes to simultaneously learn a feature extractor along with a per-class mixture component 
%(and automatically determining the number of components) 
in an online, two-phase fashion. This results in a more discriminative feature representation yielding to superior performance when applied to the few-shot image classification scenario. 
Experiments demonstrate that our approach achieves state-of-the-art results with no ancillary data used. In addition, combining our MixtFSL with \cite{Afrasiyabi_2020_ECCV} and \cite{xu2021learning}  results in significant improvements over the state of the art for inductive few-shot image classification.
A limitation of our MixtFSL is the use of a two-stage training, requiring a choreography of steps for achieving strong results while possibly increasing training time. A future line of work would be to revise it into a single stage training procedure to marry representation and mixture learning, with stable instance assignment to components, hopefully giving rise to a faster and simpler mixture model learning. Another limitation is observed with small datasets where the within-class diversity is low such that the need for mixtures is less acute (cf. CUB dataset in fig.~\ref{tab:cub}). Again, with a single-stage training, dealing with such a unimodal dataset may be better, allowing to activate multimodal mixtures only as required.

\paragraph{Acknowledgements} 
\footnotesize{
This project was supported by NSERC, Mitacs, Prompt-Quebec, and Compute Canada. %, and E Machine Learning. 
We thank M. Tremblay, H. Weber, A. Schwerdtfeger, A. Tupper, C. Shui, C. Bouchard, and B. Leger for proofreading the manuscript, and F. N. Nokabadi for help.}

{\small{ 
\bibliographystyle{ieee_fullname}
\bibliography{egbib}

\begin{thebibliography}{10}\itemsep=-1pt

\bibitem{Afrasiyabi_2020_ECCV}
Arman Afrasiyabi, Jean-Fran\c{c}ois Lalonde, and Christian Gagn\'{e}.
\newblock Associative alignment for few-shot image classification.
\newblock In {\em European Conference on Computer Vision}, 2020.

\bibitem{allen2019infinite}
Kelsey~R Allen, Evan Shelhamer, Hanul Shin, and Joshua~B Tenenbaum.
\newblock Infinite mixture prototypes for few-shot learning.
\newblock {\em International Conference on Machine Learning, PMLR}, 2019.

\bibitem{bertinetto2018metalearning}
Luca Bertinetto, Joao~F. Henriques, Philip Torr, and Andrea Vedaldi.
\newblock Meta-learning with differentiable closed-form solvers.
\newblock In {\em International Conference on Learning Representations}, 2019.

\bibitem{boudiaf2020transductive}
Malik Boudiaf, Ziko~Imtiaz Masud, J{\'e}r{\^o}me Rony, Jos{\'e} Dolz, Pablo
  Piantanida, and Ismail~Ben Ayed.
\newblock Transductive information maximization for few-shot learning.
\newblock {\em Neural Information Processing Systems}, 2020.

\bibitem{Cai_2018_CVPR}
Qi Cai, Yingwei Pan, Ting Yao, Chenggang Yan, and Tao Mei.
\newblock Memory matching networks for one-shot image recognition.
\newblock In {\em Conference on Computer Vision and Pattern Recognition}, 2018.

\bibitem{Cao_2020_CVPR}
Kaidi Cao, Jingwei Ji, Zhangjie Cao, Chien-Yi Chang, and Juan~Carlos Niebles.
\newblock Few-shot video classification via temporal alignment.
\newblock In {\em International Conference on Computer Vision and Pattern
  Recognition}, 2020.

\bibitem{caron2020unsupervised}
Mathilde Caron, Ishan Misra, Julien Mairal, Priya Goyal, Piotr Bojanowski, and
  Armand Joulin.
\newblock Unsupervised learning of visual features by contrasting cluster
  assignments.
\newblock 2020.

\bibitem{chen2019closer}
Wei-Yu Chen, Yen-Cheng Liu, Zsolt Kira, Yu-Chiang~Frank Wang, and Jia-Bin
  Huang.
\newblock A closer look at few-shot classification.
\newblock {\em arXiv preprint arXiv:1904.04232}, 2019.

\bibitem{Chen_2019_CVPR}
Zitian Chen, Yanwei Fu, Yu-Xiong Wang, Lin Ma, Wei Liu, and Martial Hebert.
\newblock Image deformation meta-networks for one-shot learning.
\newblock In {\em International Conference on Computer Vision and Pattern
  Recognition}, 2019.

\bibitem{Chu_2019_CVPR}
Wen-Hsuan Chu, Yu-Jhe Li, Jing-Cheng Chang, and Yu-Chiang~Frank Wang.
\newblock Spot and learn: A maximum-entropy patch sampler for few-shot image
  classification.
\newblock In {\em International Conference on Computer Vision and Pattern
  Recognition}, 2019.

\bibitem{deng2018arcface}
Jiankang Deng, Jia Guo, Niannan Xue, and Stefanos Zafeiriou.
\newblock Arcface: Additive angular margin loss for deep face recognition.
\newblock In {\em International Conference on Computer Vision and Pattern
  Recognition}, 2019.

\bibitem{dhillon2019baseline}
Guneet~S Dhillon, Pratik Chaudhari, Avinash Ravichandran, and Stefano Soatto.
\newblock A baseline for few-shot image classification.
\newblock {\em arXiv preprint arXiv:1909.02729}, 2019.

\bibitem{dvornik2019diversity}
Nikita Dvornik, Cordelia Schmid, and Julien Mairal.
\newblock Diversity with cooperation: Ensemble methods for few-shot
  classification.
\newblock In {\em International Conference on Computer Vision}, 2019.

\bibitem{Fan1_2020_CVPR}
Qi Fan, Wei Zhuo, Chi-Keung Tang, and Yu-Wing Tai.
\newblock Few-shot object detection with attention-rpn and multi-relation
  detector.
\newblock In {\em International Conference on Computer Vision and Pattern
  Recognition}, 2020.

\bibitem{fei2020melr}
Nanyi Fei, Zhiwu Lu, Tao Xiang, and Songfang Huang.
\newblock Melr: Meta-learning via modeling episode-level relationships for
  few-shot learning.
\newblock In {\em International Conference on Learning Representations}, 2021.

\bibitem{fei2006one}
Li Fei-Fei, Rob Fergus, and Pietro Perona.
\newblock One-shot learning of object categories.
\newblock {\em Transactions on Pattern Analysis and Machine Intelligence},
  2006.

\bibitem{fernando2012supervised}
Basura Fernando, Elisa Fromont, Damien Muselet, and Marc Sebban.
\newblock Supervised learning of gaussian mixture models for visual vocabulary
  generation.
\newblock {\em Pattern Recognition}, 2012.

\bibitem{finn2017model}
Chelsea Finn, Pieter Abbeel, and Sergey Levine.
\newblock Model-agnostic meta-learning for fast adaptation of deep networks.
\newblock In {\em International Conference on Machine Learning}, 2017.

\bibitem{finn2018probabilistic}
Chelsea Finn, Kelvin Xu, and Sergey Levine.
\newblock Probabilistic model-agnostic meta-learning.
\newblock In {\em Neural Information Processing Systems}, 2018.

\bibitem{gao2018low}
Hang Gao, Zheng Shou, Alireza Zareian, Hanwang Zhang, and Shih-Fu Chang.
\newblock Low-shot learning via covariance-preserving adversarial augmentation
  networks.
\newblock In {\em Neural Information Processing Systems}, 2018.

\bibitem{garcia2017few}
Victor Garcia and Joan Bruna.
\newblock Few-shot learning with graph neural networks.
\newblock {\em arXiv preprint arXiv:1711.04043}, 2017.

\bibitem{gauvain1994maximum}
J-L Gauvain and Chin-Hui Lee.
\newblock Maximum a posteriori estimation for multivariate gaussian mixture
  observations of markov chains.
\newblock {\em transactions on speech and audio processing}, 1994.

\bibitem{gidaris2019boosting}
Spyros Gidaris, Andrei Bursuc, Nikos Komodakis, Patrick P{\'e}rez, and Matthieu
  Cord.
\newblock Boosting few-shot visual learning with self-supervision.
\newblock In {\em International Conference on Computer Vision}, 2019.

\bibitem{gidaris2018dynamic}
Spyros Gidaris and Nikos Komodakis.
\newblock Dynamic few-shot visual learning without forgetting.
\newblock In {\em International Conference on Computer Vision and Pattern
  Recognition}, 2018.

\bibitem{gidaris2019generating}
Spyros Gidaris and Nikos Komodakis.
\newblock Generating classification weights with gnn denoising autoencoders for
  few-shot learning.
\newblock {\em arXiv preprint arXiv:1905.01102}, 2019.

\bibitem{glorot2010understanding}
Xavier Glorot and Yoshua Bengio.
\newblock Understanding the difficulty of training deep feedforward neural
  networks.
\newblock In {\em Thirteenth international conference on artificial
  intelligence and statistics}, 2010.

\bibitem{hariharan2017low}
Bharath Hariharan and Ross Girshick.
\newblock Low-shot visual recognition by shrinking and hallucinating features.
\newblock In {\em International Conference on Computer Vision}, 2017.

\bibitem{he2016deep}
Kaiming He, Xiangyu Zhang, Shaoqing Ren, and Jian Sun.
\newblock Deep residual learning for image recognition.
\newblock In {\em International Conference on Computer Vision and Pattern
  Recognition}, 2016.

\bibitem{hjort2010bayesian}
Nils~Lid Hjort, Chris Holmes, Peter M{\"u}ller, and Stephen~G Walker.
\newblock {\em Bayesian nonparametrics}.
\newblock 2010.

\bibitem{NEURIPS2019_01894d6f}
Ruibing Hou, Hong Chang, Bingpeng MA, Shiguang Shan, and Xilin Chen.
\newblock Cross attention network for few-shot classification.
\newblock In {\em Neural Information Processing Systems}, 2019.

\bibitem{hu2020empirical}
Shell~Xu Hu, Pablo~G Moreno, Yang Xiao, Xi Shen, Guillaume Obozinski, Neil~D
  Lawrence, and Andreas Damianou.
\newblock Empirical bayes transductive meta-learning with synthetic gradients.
\newblock {\em International Conference on Learning Representations}, 2020.

\bibitem{Liu_2020_ECCV}
Liu Jinlu, Song Liang, and Qin Yongqiang.
\newblock Prototype rectification for few-shot learning.
\newblock In {\em European Conference on Computer Vision}, 2020.

\bibitem{kim2019edge}
Jongmin Kim, Taesup Kim, Sungwoong Kim, and Chang~D Yoo.
\newblock Edge-labeling graph neural network for few-shot learning.
\newblock In {\em International Conference on Computer Vision and Pattern
  Recognition}, 2019.

\bibitem{kim2019variational}
Junsik Kim, Tae-Hyun Oh, Seokju Lee, Fei Pan, and In~So Kweon.
\newblock Variational prototyping-encoder: One-shot learning with prototypical
  images.
\newblock In {\em International Conference on Computer Vision and Pattern
  Recognition}, 2019.

\bibitem{kim2018bayesian}
Taesup Kim, Jaesik Yoon, Ousmane Dia, Sungwoong Kim, Yoshua Bengio, and Sungjin
  Ahn.
\newblock Bayesian model-agnostic meta-learning.
\newblock {\em arXiv preprint arXiv:1806.03836}, 2018.

\bibitem{kulis2012revisiting}
Brian Kulis and Michael~I Jordan.
\newblock Revisiting k-means: New algorithms via bayesian nonparametrics.
\newblock In {\em International Conference on Machine Learning}, 2012.

\bibitem{lee2019meta}
Kwonjoon Lee, Subhransu Maji, Avinash Ravichandran, and Stefano Soatto.
\newblock Meta-learning with differentiable convex optimization.
\newblock In {\em International Conference on Computer Vision and Pattern
  Recognition}, 2019.

\bibitem{li2021learning}
Junjie Li, Zilei Wang, and Xiaoming Hu.
\newblock Learning intact features by erasing-inpainting for few-shot
  classification.
\newblock In {\em Proceedings of the AAAI Conference on Artificial
  Intelligence}, volume~35, 2021.

\bibitem{Li_2019_CVPR}
Wenbin Li, Lei Wang, Jinglin Xu, Jing Huo, Yang Gao, and Jiebo Luo.
\newblock Revisiting local descriptor based image-to-class measure for few-shot
  learning.
\newblock In {\em International Conference on Computer Vision and Pattern
  Recognition}, 2019.

\bibitem{lifchitz2019dense}
Yann Lifchitz, Yannis Avrithis, Sylvaine Picard, and Andrei Bursuc.
\newblock Dense classification and implanting for few-shot learning.
\newblock In {\em International Conference on Computer Vision and Pattern
  Recognition}, 2019.

\bibitem{Bin_2020_ECCV_margin_matter}
Bin Liu, Yue Cao, Yutong Lin, Qi Li, Zheng Zhang, Mingsheng Long, and Han Hu.
\newblock Negative margin matters: Understanding margin in few-shot
  classification.
\newblock In {\em European Conference on Computer Vision}, 2020.

\bibitem{Liu_2019_ICCV}
Bin Liu, Zhirong Wu, Han Hu, and Stephen Lin.
\newblock Deep metric transfer for label propagation with limited annotated
  data.
\newblock In {\em International Conference on Computer Vision Workshops}, 2019.

\bibitem{liu2018learning}
Yanbin Liu, Juho Lee, Minseop Park, Saehoon Kim, Eunho Yang, Sung~Ju Hwang, and
  Yi Yang.
\newblock Learning to propagate labels: Transductive propagation network for
  few-shot learning.
\newblock {\em arXiv preprint arXiv:1805.10002}, 2018.

\bibitem{maaten2008visualizing}
Laurens van~der Maaten and Geoffrey Hinton.
\newblock Visualizing data using t-sne.
\newblock {\em Journal of Machine Learning Research}, 2008.

\bibitem{mehrotra2017generative}
Akshay Mehrotra and Ambedkar Dukkipati.
\newblock Generative adversarial residual pairwise networks for one shot
  learning.
\newblock {\em arXiv preprint arXiv:1703.08033}, 2017.

\bibitem{mishra2017simple}
Nikhil Mishra, Mostafa Rohaninejad, Xi Chen, and Pieter Abbeel.
\newblock A simple neural attentive meta-learner.
\newblock {\em arXiv preprint arXiv:1707.03141}, 2017.

\bibitem{mnih2015human}
Volodymyr Mnih, Koray Kavukcuoglu, David Silver, Andrei~A Rusu, Joel Veness,
  Marc~G Bellemare, Alex Graves, Martin Riedmiller, Andreas~K Fidjeland, Georg
  Ostrovski, et~al.
\newblock Human-level control through deep reinforcement learning.
\newblock {\em nature}, 2015.

\bibitem{munkhdalai2018rapid}
Tsendsuren Munkhdalai, Xingdi Yuan, Soroush Mehri, and Adam Trischler.
\newblock Rapid adaptation with conditionally shifted neurons.
\newblock In {\em International Conference on Machine Learning}, 2018.

\bibitem{oreshkin2018tadam}
Boris Oreshkin, Pau~Rodr{\'\i}guez L{\'o}pez, and Alexandre Lacoste.
\newblock Tadam: Task dependent adaptive metric for improved few-shot learning.
\newblock In {\em Neural Information Processing Systems}, 2018.

\bibitem{Perez-Rua_2020_CVPR}
Juan-Manuel Perez-Rua, Xiatian Zhu, Timothy~M. Hospedales, and Tao Xiang.
\newblock Incremental few-shot object detection.
\newblock In {\em International Conference on Computer Vision and Pattern
  Recognition}, 2020.

\bibitem{Qi_2018_CVPR}
Hang Qi, Matthew Brown, and David~G. Lowe.
\newblock Low-shot learning with imprinted weights.
\newblock In {\em International Conference on Computer Vision and Pattern
  Recognition}, 2018.

\bibitem{Qiao_2019_ICCV}
Limeng Qiao, Yemin Shi, Jia Li, Yaowei Wang, Tiejun Huang, and Yonghong Tian.
\newblock Transductive episodic-wise adaptive metric for few-shot learning.
\newblock In {\em IEEE/CVF International Conference on Computer Vision}, 2019.

\bibitem{Qiao_2018_CVPR}
Siyuan Qiao, Chenxi Liu, Wei Shen, and Alan~L. Yuille.
\newblock Few-shot image recognition by predicting parameters from activations.
\newblock In {\em International Conference on Computer Vision and Pattern
  Recognition}, 2018.

\bibitem{rasmussen2000infinite}
Carl~Edward Rasmussen.
\newblock The infinite gaussian mixture model.
\newblock In {\em Neural Information Processing Systems}, 2000.

\bibitem{ravi2016optimization}
Sachin Ravi and Hugo Larochelle.
\newblock Optimization as a model for few-shot learning.
\newblock In {\em International Conference on Learning Representations}, 2016.

\bibitem{NEURIPS2019_5f8e2fa1}
Ali Razavi, Aaron van~den Oord, and Oriol Vinyals.
\newblock Generating diverse high-fidelity images with vq-vae-2.
\newblock In {\em Neural Information Processing Systems}, 2019.

\bibitem{ren2018meta}
Mengye Ren, Eleni Triantafillou, Sachin Ravi, Jake Snell, Kevin Swersky,
  Joshua~B Tenenbaum, Hugo Larochelle, and Richard~S Zemel.
\newblock Meta-learning for semi-supervised few-shot classification.
\newblock {\em arXiv preprint arXiv:1803.00676}, 2018.

\bibitem{russakovsky2015imagenet}
Olga Russakovsky, Jia Deng, Hao Su, Jonathan Krause, Sanjeev Satheesh, Sean Ma,
  Zhiheng Huang, Andrej Karpathy, Aditya Khosla, Michael Bernstein, et~al.
\newblock Imagenet large scale visual recognition challenge.
\newblock {\em International Journal of Computer Vision}, 2015.

\bibitem{rusu2018meta}
Andrei~A Rusu, Dushyant Rao, Jakub Sygnowski, Oriol Vinyals, Razvan Pascanu,
  Simon Osindero, and Raia Hadsell.
\newblock Meta-learning with latent embedding optimization.
\newblock {\em International Conference on Learning Representations}, 2018.

\bibitem{schwartz2018delta}
Eli Schwartz, Leonid Karlinsky, Joseph Shtok, Sivan Harary, Mattias Marder,
  Abhishek Kumar, Rogerio Feris, Raja Giryes, and Alex Bronstein.
\newblock Delta-encoder: an effective sample synthesis method for few-shot
  object recognition.
\newblock In {\em Neural Information Processing Systems}, 2018.

\bibitem{sergey2016wide}
Zagoruyko Sergey and Komodakis Nikos.
\newblock Wide residual networks.
\newblock In {\em British Machine Vision Conference}, 2016.

\bibitem{Simon_2020_CVPR}
Christian Simon, Piotr Koniusz, Richard Nock, and Mehrtash Harandi.
\newblock Adaptive subspaces for few-shot learning.
\newblock In {\em International Conference on Computer Vision and Pattern
  Recognition}, 2020.

\bibitem{simon2020modulating}
Christian Simon, Piotr Koniusz, Richard Nock, and Mehrtash Harandi.
\newblock On modulating the gradient for meta-learning.
\newblock In {\em European Conference on Computer Vision}, 2020.

\bibitem{snell2017prototypical}
Jake Snell, Kevin Swersky, and Richard Zemel.
\newblock Prototypical networks for few-shot learning.
\newblock In {\em Neural Information Processing Systems}, 2017.

\bibitem{su2020does}
Jong-Chyi Su, Subhransu Maji, and Bharath Hariharan.
\newblock When does self-supervision improve few-shot learning?
\newblock In {\em European Conference on Computer Vision}, 2020.

\bibitem{sun2019meta}
Qianru Sun, Yaoyao Liu, Tat-Seng Chua, and Bernt Schiele.
\newblock Meta-transfer learning for few-shot learning.
\newblock In {\em International Conference on Computer Vision and Pattern
  Recognition}, 2019.

\bibitem{sung2018learning}
Flood Sung, Yongxin Yang, Li Zhang, Tao Xiang, Philip~H.S. Torr, and Timothy~M.
  Hospedales.
\newblock Learning to compare: Relation network for few-shot learning.
\newblock In {\em International Conference on Computer Vision and Pattern
  Recognition}, 2018.

\bibitem{tang2020long}
Kaihua Tang, Jianqiang Huang, and Hanwang Zhang.
\newblock Long-tailed classification by keeping the good and removing the bad
  momentum causal effect.
\newblock {\em Neural Information Processing Systems}, 2020.

\bibitem{Tian_2020_ECCV_good}
Yonglong Tian, Yue Wang, Dilip Krishnan, Joshua~B Tenenbaum, and Phillip Isola.
\newblock Few-shot image classification: a good embedding is all you need.
\newblock In {\em European Conference on Computer Vision}, 2020.

\bibitem{tseng2020cross}
Hung-Yu Tseng, Hsin-Ying Lee, Jia-Bin Huang, and Ming-Hsuan Yang.
\newblock Cross-domain few-shot classification via learned feature-wise
  transformation.
\newblock {\em arXiv preprint arXiv:2001.08735}, 2020.

\bibitem{van2017neural}
Aaron Van Den~Oord, Oriol Vinyals, et~al.
\newblock Neural discrete representation learning.
\newblock In {\em Neural Information Processing Systems}, 2017.

\bibitem{vilalta2002perspective}
Ricardo Vilalta and Youssef Drissi.
\newblock A perspective view and survey of meta-learning.
\newblock {\em Artificial Intelligence Review}, 2002.

\bibitem{vinyals2016matching}
Oriol Vinyals, Charles Blundell, Timothy Lillicrap, Daan Wierstra, et~al.
\newblock Matching networks for one shot learning.
\newblock In {\em Neural Information Processing Systems}, 2016.

\bibitem{wah2011caltech}
Catherine Wah, Steve Branson, Peter Welinder, Pietro Perona, and Serge
  Belongie.
\newblock The {Caltech-UCSD} birds-200-2011 dataset, 2011.

\bibitem{Wang3_2020_CVPR}
Yaxing Wang, Salman Khan, Abel Gonzalez-Garcia, Joost van~de Weijer, and
  Fahad~Shahbaz Khan.
\newblock Semi-supervised learning for few-shot image-to-image translation.
\newblock In {\em International Conference on Computer Vision and Pattern
  Recognition}, 2020.

\bibitem{Wang1_2020_CVPR}
Yikai Wang, Chengming Xu, Chen Liu, Li Zhang, and Yanwei Fu.
\newblock Instance credibility inference for few-shot learning.
\newblock In {\em International Conference on Computer Vision and Pattern
  Recognition}, 2020.

\bibitem{wang2018low}
Yu-Xiong Wang, Ross Girshick, Martial Hebert, and Bharath Hariharan.
\newblock Low-shot learning from imaginary data.
\newblock In {\em International Conference on Computer Vision and Pattern
  Recognition}, 2018.

\bibitem{wang2016learning}
Yu-Xiong Wang and Martial Hebert.
\newblock Learning from small sample sets by combining unsupervised
  meta-training with cnns.
\newblock In {\em Neural Information Processing Systems}, 2016.

\bibitem{wang2019meta}
Yu-Xiong Wang, Deva Ramanan, and Martial Hebert.
\newblock Meta-learning to detect rare objects.
\newblock In {\em International Conference on Computer Vision}, 2019.

\bibitem{wertheimer2019few}
Davis Wertheimer and Bharath Hariharan.
\newblock Few-shot learning with localization in realistic settings.
\newblock In {\em International Conference on Computer Vision and Pattern
  Recognition}, 2019.

\bibitem{west1993hierarchical}
Mike West and Michael~D Escobar.
\newblock {\em Hierarchical priors and mixture models, with application in
  regression and density estimation}.
\newblock 1993.

\bibitem{xu2021learning}
Chengming Xu, Yanwei Fu, Chen Liu, Chengjie Wang, Jilin Li, Feiyue Huang, Li
  Zhang, and Xiangyang Xue.
\newblock Learning dynamic alignment via meta-filter for few-shot learning.
\newblock In {\em Conference on Computer Vision and Pattern Recognition}, 2021.

\bibitem{yoon2019tapnet}
Sung~Whan Yoon, Jun Seo, and Jaekyun Moon.
\newblock Tapnet: Neural network augmented with task-adaptive projection for
  few-shot learning.
\newblock {\em International Conference on Machine Learning}, 2019.

\bibitem{Zhang_2020_CVPR}
Chi Zhang, Yujun Cai, Guosheng Lin, and Chunhua Shen.
\newblock Deepemd: Few-shot image classification with differentiable earth
  mover's distance and structured classifiers.
\newblock In {\em International Conference on Computer Vision and Pattern
  Recognition}, 2020.

\bibitem{zhang2017mixup}
Hongyi Zhang, Moustapha Cisse, Yann~N Dauphin, and David Lopez-Paz.
\newblock mixup: Beyond empirical risk minimization.
\newblock {\em arXiv preprint arXiv:1710.09412}, 2017.

\bibitem{Zhang_2019_CVPR}
Hongguang Zhang, Jing Zhang, and Piotr Koniusz.
\newblock Few-shot learning via saliency-guided hallucination of samples.
\newblock In {\em International Conference on Computer Vision and Pattern
  Recognition}, 2019.

\bibitem{zhang2019variational}
Jian Zhang, Chenglong Zhao, Bingbing Ni, Minghao Xu, and Xiaokang Yang.
\newblock Variational few-shot learning.
\newblock In {\em International Conference on Computer Vision}, 2019.

\bibitem{zhang2020iept}
Manli Zhang, Jianhong Zhang, Zhiwu Lu, Tao Xiang, Mingyu Ding, and Songfang
  Huang.
\newblock Iept: Instance-level and episode-level pretext tasks for few-shot
  learning.
\newblock In {\em International Conference on Learning Representations}, 2021.

\bibitem{zhao2021looking}
Jiabao Zhao, Yifan Yang, Xin Lin, Jing Yang, and Liang He.
\newblock Looking wider for better adaptive representation in few-shot
  learning.
\newblock In {\em Proceedings of the AAAI Conference on Artificial
  Intelligence}, 2021.

\bibitem{masud2020laplacian}
Imtiaz Ziko, Jose Dolz, Eric Granger, and Ismail~Ben Ayed.
\newblock Laplacian regularized few-shot learning.
\newblock In {\em International Conference on Machine Learning}, 2020.

\end{thebibliography}
}}

\onecolumn 
%%%%%%%%% TITLE 
\begin{center}\Large
    Mixture-based Feature Space Learning for Few-shot Image Classification\\ Supplementary Material
\end{center}

\ \\ \ \\ \ \\ \ \\ \ \\ \ \\ 
 
In this supplementary material, the following items are provided:
\begin{enumerate} 

    \item Ablation on the number of components $N^k$ in the mixture model $\bm{\mathcal{P}}$ (sec.~\ref{sec:mixture_model_size})
    \item Dynamic of the training (sec.~\ref{sec:Dynamic});
    
    \item More ways ablation (sec.~\ref{sec:number_of_ways});

    \item Ablation of the margin $m$ (sec.~\ref{sec:effect_of_margin}); 
    \item Ablation of the temperature $\tau$ (sec.~\ref{sec:effect_of_temperature});

    \item Visualization: from MixtFSL to MixtFSL-Alignment (sec.~\ref{sec:visualization});

\end{enumerate} 
    
%\end{abstract}

\ \\ \ \\ \ \\ \ \\ \ \\

\newpage
\section{Ablation on the number of components $N^k$ in the mixture model $\bm{\mathcal{P}}$}
\label{sec:mixture_model_size}

% \begin{table*}[hbt!]
% \centering
% \caption{N-way classification results on mini-ImageNet using ResNet-12 and ResNet-18 backbones.   $\pm$ denotes the $95\%$ confidence intervals over 600 episodes.} 
%     \begin{tabular}{lcccccccccc}  % rrccccc
%     \toprule
%       &  & \multicolumn{2}{c}{\textbf{ResNet-12}}  
%       & & \multicolumn{2}{c}{ \textbf{ResNet-18}}    
%         \\ 
        
%         & \textbf{$N^k$}    \quad   \quad                 
%         & 1-shot  & 5-shot  
%         &
%         & 1-shot  & 5-shot  
%         \\ 
%         \midrule 
        
%         &$5$  \quad   \quad       
%         &62.29\scriptsize{ $\pm$1.08}       &78.85\scriptsize{ $\pm$0.61} 
%         &
%         &58.57\scriptsize{ $\pm$1.09}       &76.44\scriptsize{ $\pm$0.61}  
%         \\
        
%         &$10$  \quad   \quad         
%         &64.01\scriptsize{ $\pm$0.79}       &81.87\scriptsize{ $\pm$0.49} 
%         &
%         &60.15\scriptsize{ $\pm$0.80}       &77.71\scriptsize{ $\pm$0.61}  
%         \\

%         &$15$    \quad   \quad     
%         &63.98\scriptsize{ $\pm$0.79}       &82.04\scriptsize{ $\pm$0.49} 
%         &
%         &60.11\scriptsize{ $\pm$0.73}      &77.76\scriptsize{ $\pm$0.58}  
%         \\

%         &$20$    \quad   \quad    
%         &63.91\scriptsize{ $\pm$0.80}       & 82.05\scriptsize{ $\pm$0.49} 
%         &
%         &58.99\scriptsize{ $\pm$0.81}       &77.73\scriptsize{ $\pm$0.58}  
%         \\

%     \end{tabular}
%     \\   
%     \label{tab:P_size}
% \end{table*}

Although our proposed MixtFSL automatically infers the number of per-class mixture components from data, we also ablate the initial size of mixture model $N^k$ for each class to evaluate whether it has an impact on the final results.  Table~\ref{tab:N_k} presents 1- and 5-shot classification results on miniImageNet using ResNet-12 and ResNet-18 by initializing $N^k$ to 5, 10, 15, and 20 components per class. 

Initializing $N^k=5$ results in lower classification accuracy compared to the higher $N^k$. We think this is possible due to the insufficient capacity of small mixture model $\bm{\mathcal{P}}$ size. 
However, as long as $N^k$ is sufficiently large (10, 15, 20), our approach is robust to this parameter and results do not change significantly as a function of $N^k$. Note that $N^k$ cannot be set to an arbitrary high number due to memory limitations.

\begin{table*}[hbt!]
\centering
\caption{Classification results on mini-ImageNet using ResNet-12 and ResNet-18 backbones as a function of the initial value for the number of components per class $N^k$. $\pm$ denotes the $95\%$ confidence intervals over 300 episodes.} 
  \begin{tabular}{cc}
    \begin{tabular}{ccc}         %& & \multicolumn{2}{c}{ResNet-18}   \\  %\cline{3-4}  \cline{6-7}    
    % adversarial
        \toprule
     
        $N^k $& 1-shot  & 5-shot \\ 
        \midrule
        5  &62.29\scriptsize{ $\pm$1.08}       &78.85\scriptsize{ $\pm$0.61} \\
        10 &64.01\scriptsize{ $\pm$0.79}       &81.87\scriptsize{ $\pm$0.49} \\
        15 &63.98\scriptsize{ $\pm$0.79}       &82.04\scriptsize{ $\pm$0.49} \\
        20 &63.91\scriptsize{ $\pm$0.80}       &82.05\scriptsize{ $\pm$0.49} \\
        \bottomrule
    \end{tabular}
    & 
    \begin{tabular}{ccc} 
    % centroid
        %& & \multicolumn{2}{c}{ResNet-18}   \\  %\cline{3-4}  \cline{6-7}    
        \toprule
        
        $N^k$ & 1-shot  & 5-shot \\ 
        \midrule
        5 &58.57\scriptsize{ $\pm$1.09}       &76.44\scriptsize{ $\pm$0.61}  \\
        10 &60.15\scriptsize{ $\pm$0.80}       &77.71\scriptsize{ $\pm$0.61}  \\
        15 &60.11\scriptsize{ $\pm$0.73}      &77.76\scriptsize{ $\pm$0.58}  \\
        20 &58.99\scriptsize{ $\pm$0.81}       &77.77\scriptsize{ $\pm$0.58}  \\
        
        \bottomrule
    \end{tabular}
    \\
    (a) ResNet-12 & (b) ResNet-18
    \end{tabular}
    
    %$^\S$ ResNet-12 with additional learning module \ \\
    %$^{\dag\diamond}$ our implementation, with early stopping \quad \ \\
    %$^\ddag$ implementation from \cite{chen2019closer}  \ \\
    % $\diamond$ ResNet-12 plus additional learning module 
    \label{tab:N_k}
\end{table*}

\newpage
\section{Dynamic of the training}
\label{sec:Dynamic}
\begin{figure}
    \centering
    \footnotesize
    \setlength{\tabcolsep}{1pt}
    \begin{tabular}{cc} 
    \includegraphics[width=0.3\linewidth]{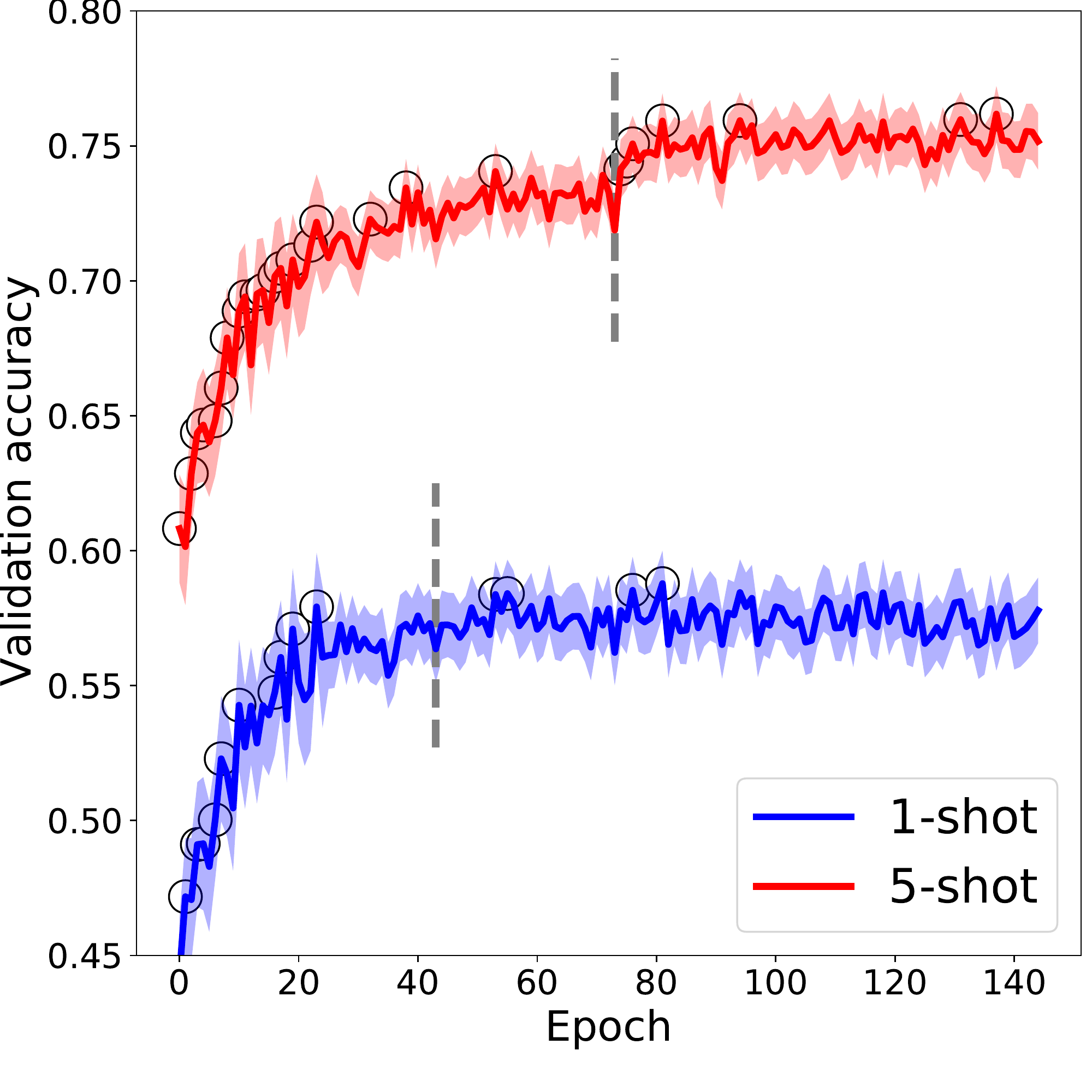} & \ \ 
    \includegraphics[width=0.3\linewidth]{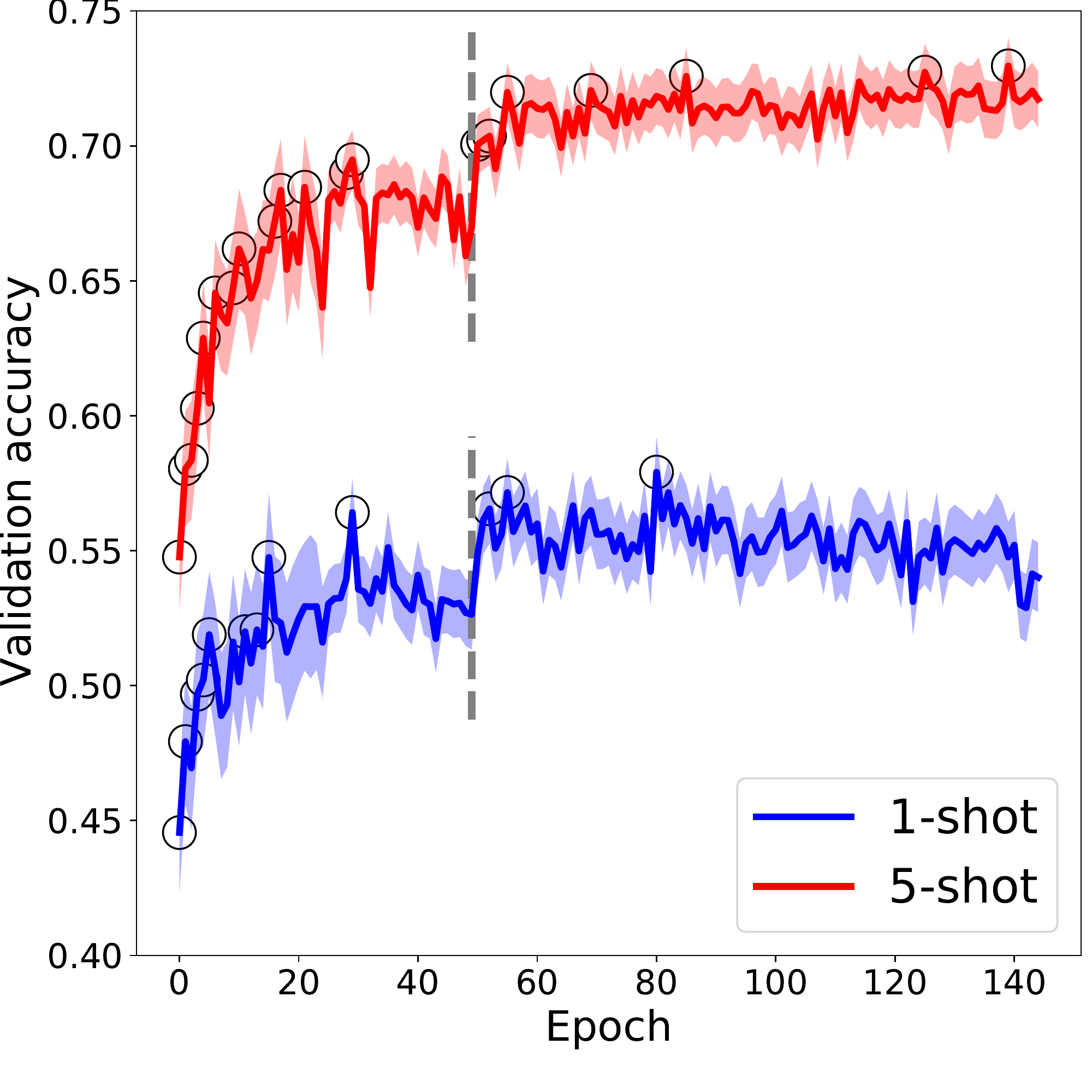} \\ 
    (a) ResNet-12  & 
    (b) ResNet-18
    \end{tabular}
    \caption{Validation accuracy of the first 150 epochs using ResNet-12 and ResNet-18 on miniImageNet. 1- and 5-shot scenarios are plotted using blue and red colors with their confidence intervals over 300 testing episodes of the validation set, respectively. The dashed vertical line is starting point of progressive following stage. The circles are the points when we update the best model.}
    \label{fig:validationAccuracy}
\end{figure}

Fig.~\ref{fig:validationAccuracy} evaluates the necessity of the two training stages (sec.~4 from the main paper) by showing the (episodic) validation accuracy during 150 epochs. The vertical dashed line indicates the transition between training stages. In most cases, the progressive following stage results in a validation accuracy gain. 

\newpage
\section{More ways ablation}
\label{sec:number_of_ways}

Table~\ref{tab:More_way} presents more-way 5-shot comparison of our MixtFSL on miniImageNet using ResNet-18 and ResNet-12. Our MixtFSL gains 1.14\% and 1.23\% over the Pos-Margin~\cite{Afrasiyabi_2020_ECCV} in 5-way and 20-way, respectively. Besides, MixtFSL gains 0.78\% over Baseline++~\cite{chen2019closer} in 10-way.  

We could not find ``more-ways'' results with the ResNet-12 backbone in the literature, but we provide our results here for potential future literature comparisons.

%\todo{Arman: couldn't find more way sing ResNet-12, but I guess it might be better to keep it and if somebody asked we can say its there}

%We experiment with N-way, 5-shot experiment (for N = 5, 10, 20) to examine the effect of associative alignment on more-way using mini-ImageNet. As Table~\ref{tab:N-way} presents, our associative alignment gains on the compared meta-learning and standard transfer learning methods. Specifically, we outperform the best of the compared method by 6.67\%, 4.47\%, 3.82\% in 5-, 10-, and 20-way respectively. Note that we used 10, 5, 3 number of related base classes (B) 5-way, 10-way and 20-way respectively which corresponds to 60 classes out of all 64 base categories in mini-ImageNet.   

\begin{table*}[hbt!]
\centering
\caption{$N$-way 5-shot classification results on mini-ImageNet using ResNet-18 and ResNet-12 backbones.   $\pm$ denotes the $95\%$ confidence intervals over 600 episodes.
The best results prior this work is highlighted in blue, and the best results are presented in boldfaced.} 
    \begin{tabular}{ rrccccccc}  % rrccccc
        \toprule

        %& & \multicolumn{2}{c}{Conv4} & & \multicolumn{2}{c}{ResNet-18}   \\  %\cline{3-4}  \cline{6-7}           
          & Method   & Backbone  & \textbf{5-way}  &  & \textbf{10-way}  &   & \textbf{20-way}            \\ % & --  & -- \\ 
        \midrule
        
    %\multirow{3}{*}{\rotatebox{90}{meta-l.}} 
        & MatchingNet$^\ddag$~\cite{vinyals2016matching}  & RN-18
        & 68.88\scriptsize{ $\pm$0.69}       &       & 52.27\scriptsize{ $\pm$0.46}     &        & 36.78\scriptsize{ $\pm$0.25}  
        \\

        & ProtoNet$^\ddag$~\cite{snell2017prototypical}    &RN-18
        & 73.68\scriptsize{ $\pm$0.65}        &         & 59.22\scriptsize{ $\pm$0.44}         &       &  44.96\scriptsize{ $\pm$0.26}             
        \\

        %\multicolumn{2}{r}{no alignment} 
        & RelationNet$^\ddag$~\cite{sung2018learning}             &RN-18  
        & 69.83\scriptsize{ $\pm$0.68}                               &           
        & 53.88\scriptsize{ $\pm$0.48}                                 &        
        & 39.17\scriptsize{ $\pm$0.25}         
        \\      
  
        & Baseline~\cite{chen2019closer}                &RN-18
        & 74.27\scriptsize{ $\pm$0.63}                   &           
        & 55.00\scriptsize{ $\pm$0.46}                  &
        & 42.03\scriptsize{ $\pm$0.25}                  
        \\

        & Baseline++~\cite{chen2019closer}              &RN-18
        & 75.68\scriptsize{ $\pm$0.63}                &        
        & {\color{blue} 63.40}\scriptsize{ $\pm$0.44}                 &
        & 50.85\scriptsize{ $\pm$0.25}                  
        \\

        & Pos-Margin~\cite{Afrasiyabi_2020_ECCV}                     &RN-18
        & {\color{blue}76.62}\scriptsize{ $\pm$0.58}                 &
        & 62.95\scriptsize{ $\pm$0.83}                &
        & {\color{blue}51.92}\scriptsize{ $\pm$1.02}                     
        \\*[0.5em]
 
        & MixtFSL (ours)                                                   &RN-18
        & \textbf{77.76}\scriptsize{ $\pm$0.58}                        &
        & \textbf{64.18}\scriptsize{ $\pm$0.76}                                 &
        & \textbf{53.15}\scriptsize{ $\pm$0.71}                     
        \\
        
        \cmidrule(lr){2-8}
        & MixtFSL (ours)                                                   &RN-12
        & 82.04\scriptsize{ $\pm$0.49}                        &
        & 68.26\scriptsize{ $\pm$0.71}                                 &
        & 55.41\scriptsize{ $\pm$0.71}                     
        \\

    \toprule
    \end{tabular}
    \\  
    $^\ddag$ implementation from \cite{chen2019closer}  
    \label{tab:More_way}
\end{table*}

 \newpage
\section{Ablation of the margin}
\label{sec:effect_of_margin}  

As table~\ref{tab:margin-ablation} shows, a negative margin provides slightly better results than using a positive one, thus replicating the findings from Liu~\etal~\cite{Bin_2020_ECCV_margin_matter}, albeit with a more modest improvement than reported in their paper. We theorize that the differences between our results (in table~\ref{tab:margin-ablation}) and theirs are due to slight differences in training setup (e.g., learning rate scheduling, same optimizer for base and novel classes). Nevertheless, the impact of the margin on our proposed  MixtFSL approach is similar. 
We also note that in all cases except 5-shot on ResNet-18, our proposed MixtFSL yields significant improvements. Notably,  MixtFSL provides classification improvements of 2.08\% and 3.18\% in 1-shot and 5-shot using ResNet-12. 

\begin{table}[t]
\renewcommand{\tabcolsep}{2pt}
\centering
\caption{Margin evaluation using miniImageNet in 5-way classification. Bold/blue is best/second best, and $\pm$ indicates the 95\% confidence intervals over 600 episodes.} 

\begin{tabular}{rlccc}  
        \toprule  
        & \textbf{Method}   
        & \textbf{\small Backbone} 
        & \textbf{1-shot}  
        & \textbf{5-shot}    
        \\  
        \midrule   
            & Neg-Margin$^*$~\cite{Bin_2020_ECCV_margin_matter}       
                & \small{Conv4} 
                & 51.81\scriptsize{ $\pm$0.81}      & {\color{blue}69.24}\scriptsize{ $\pm$0.59} 
            \\
            % & Pos-Margin$^*$~\cite{Afrasiyabi_2020_ECCV}
            & ArcMax$^*$~\cite{Afrasiyabi_2020_ECCV}  
                & \small{Conv4} 
                & {\color{blue}51.95}\scriptsize{ $\pm$0.80}      & 69.05\scriptsize{ $\pm$ 0.58} 
         \\*[0.5em]

         %----------------------------------MixtFSL(Conv4)------------------------------------% 
            & MixtFSL-Neg-Margin
                & \small{Conv4} 
                & 52.76\scriptsize{ $\pm$0.67}      & \textbf{70.67}\scriptsize{ $\pm$0.57} 
            \\
            & MixtFSL-Pos-Margin     
                & \small{Conv4} 
                & \textbf{52.82}\scriptsize{ $\pm$0.63}       & 70.30\scriptsize{ $\pm$0.59} 
            \\%*[0.5em]   
            
        \midrule
        %----------------------------------(RN12)------------------------------------% 
            & Neg-Margin$^*$~\cite{Bin_2020_ECCV_margin_matter}      
                & \small{RN-12} 
                & {\color{blue}61.90}\scriptsize{ $\pm$0.74}      & {\color{blue}78.86}\scriptsize{ $\pm$0.53}      
         \\
         
            % & Pos-Margin$^*$~\cite{Afrasiyabi_2020_ECCV}
            & ArcMax$^*$~\cite{Afrasiyabi_2020_ECCV}  
                & \small{RN-12} 
                & 61.86\scriptsize{ $\pm$0.71}      & 78.55\scriptsize{ $\pm$0.55} 
            \\*[0.5em]

         %----------------------------------MixtFSL(RN12)------------------------------------% 
            & MixtFSL-Neg-Margin    
                & \small{RN-12} 
                & \textbf{63.98}\scriptsize{ $\pm$0.79}       & \textbf{82.04}\scriptsize{ $\pm$0.49} 
            \\
            &  MixtFSL-Pos-Margin    
                & \small{RN-12} 
                & 63.57\scriptsize{ $\pm$0.00}      & 81.70\scriptsize{ $\pm$0.49} 
            \\%*[0.5em]   
            
        \midrule      
        % %----------------------------------(RN18)------------------------------------% 
            & Neg-Margin$^*$~\cite{Bin_2020_ECCV_margin_matter}    
                & \small{RN-18} 
                & {\color{blue}59.15}\scriptsize{ $\pm$0.81}      & \textbf{78.41}\scriptsize{ $\pm$0.54}       
            \\  
        %   & Pos-Margin$^*$~\cite{Afrasiyabi_2020_ECCV}
        & ArcMax$^*$~\cite{Afrasiyabi_2020_ECCV}  
                & \small{RN-18} 
                & 58.42\scriptsize{ $\pm$0.84}      & 77.72\scriptsize{ $\pm$0.51}   
            \\*[0.5em]
        %----------------------------------MixtFSL(RN18)------------------------------------%
            & MixtFSL-Neg-Margin
                & \small{RN-18} 
                & \textbf{60.11}\scriptsize{ $\pm$0.73}      & {\color{blue}{77.76}}\scriptsize{ $\pm$0.58}
                \\  
            & MixtFSL-Pos-Margin        
                & \small{RN-18} 
                & 59.71\scriptsize{ $\pm$0.76}     & {77.59}\scriptsize{ $\pm$0.58} 
                \\%*[0.5em]    

          \midrule   
        %----------------------------------(RN18)------------------------------------% 
        & Neg-Margin$^*$~\cite{Bin_2020_ECCV_margin_matter}       
                & \small{WRN} 
                & 62.27\scriptsize{ $\pm$0.90}      & 80.52\scriptsize{ $\pm$0.49}    
        \\
        % & Pos-Margin$^*$~\cite{Afrasiyabi_2020_ECCV}
        & ArcMax$^*$~\cite{Afrasiyabi_2020_ECCV}  
                & \small{WRN} 
                & {\color{blue}62.68}\scriptsize{ $\pm$0.76}      & {\color{blue}80.54}\scriptsize{ $\pm$0.50}   
            \\*[0.5em]  
        %----------------------------------MixtFSL(RN18)------------------------------------%
            & MixtFSL-Neg-Margin
                & \small{WRN} 
                & 63.18\scriptsize{ $\pm$1.02}      &81.66\scriptsize{ $\pm$0.60}
                \\  
            & MixtFSL-Pos-Margin        
                & \small{WRN} 
                &  \textbf{64.31}\scriptsize{ $\pm$0.79}     & \textbf{81.63}\scriptsize{ $\pm$0.56} 
                \\%*[0.5em]   
         \bottomrule 
        
    \end{tabular}  \\  
    \label{tab:margin-ablation} 
    {\footnotesize $^*$ our implementation}  
\end{table}
The margin $m$ in eq.1 (sec. 4.1) is ablated in Table~\ref{tab:margin_ablation_MixtFSL} using the validation set of the miniImagNet dataset using ResNet-12 and ResNet-18. We experiment with both $m=0.01$ to match Afrasiyabi~\etal~\cite{Afrasiyabi_2020_ECCV}, and $m=-0.02$ to match Bin~\etal~\cite{Bin_2020_ECCV_margin_matter}.

\begin{table*}[h!]
\centering
\caption{Margin $m$ ablation on the miniImageNet using ResNet-12 and ResNet-18 backbones.  }%$\pm$ denotes the $95\%$ confidence intervals over 300 episodes.} 
    \begin{tabular}{ rrcccccc}   
      \toprule
        & & & \multicolumn{2}{c}{ResNet-12} & & \multicolumn{2}{c}{ResNet-18}   \\      
        &  & $m$                           & 1-shot  & 5-shot & & 1-shot  & 5-shot \\ 
        \midrule 
        \multicolumn{2}{r}{} &-0.02       & 61.85    & 80.38  & & 60.57   & 79.04          \\
        \multicolumn{2}{r}{} &+0.01       & 60.97    & 77.43  & &60.27   & 78.12       \\  
        \bottomrule
    \end{tabular}
    \\ 
    \label{tab:margin_ablation_MixtFSL}
\end{table*}

% We used episodic cross-validation to find the margin ($m$). In our experiments, we found that $m$ needs to be adjusted according to the architectures rather than the datasets, which is likely due to its relation to the network learning capacity. An ablation for $m$ on the \emph{mini}-ImageNet validation set for the 5-way scenario is presented in table~\ref{tab:margin_ablation}.

\newpage
\section{Ablation of the temperature $\tau$}
\label{sec:effect_of_temperature} 

\begin{figure}[!h]
    \centering
    \includegraphics[height=5cm]{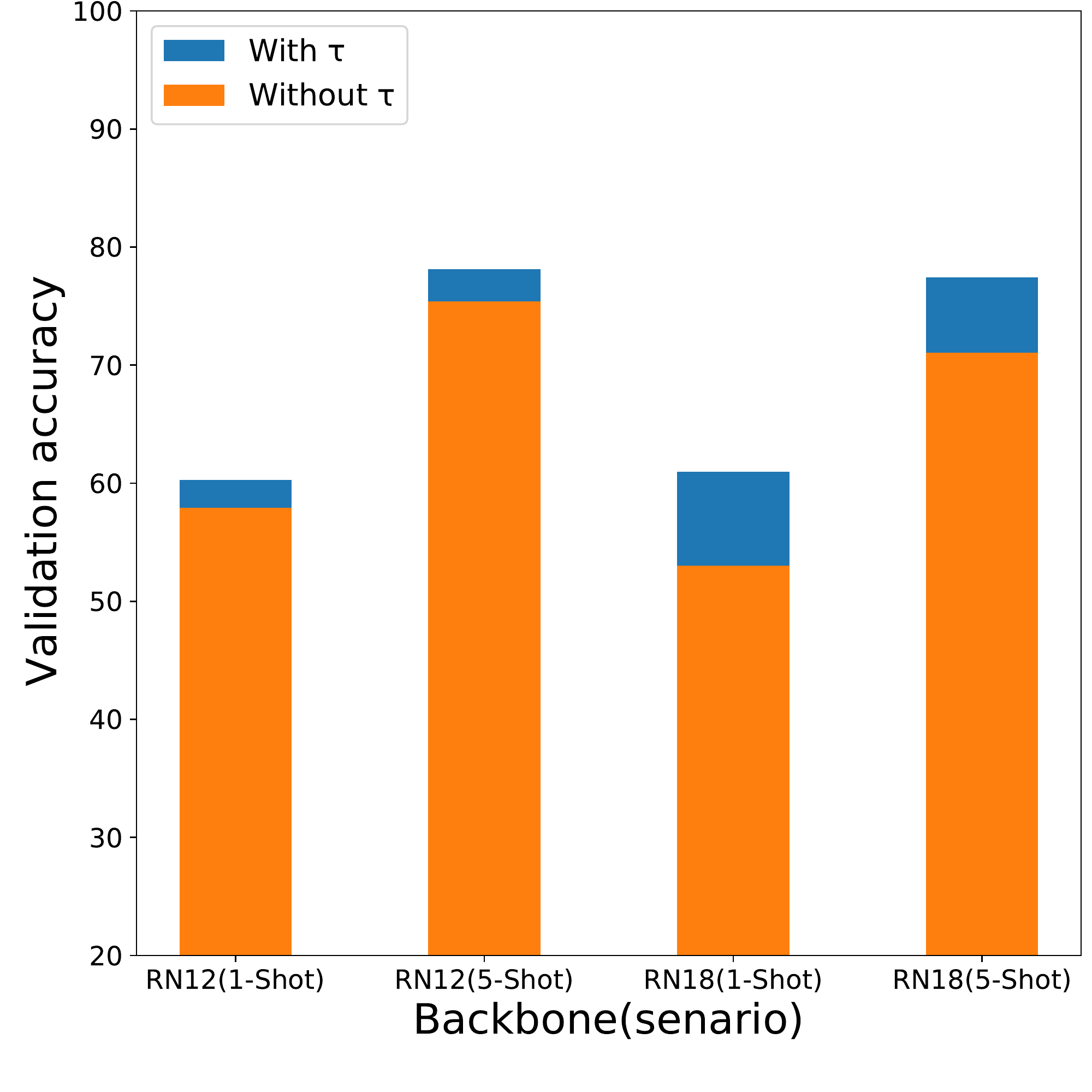}
    \caption{Effect of temperature $\tau$ on MixtFSL using ResNet-12 and -18 in 1- and 5-shot scenarios in miniImageNet's validation set. The orange bars are the classification results without temperature variable ($\tau=1$), and the blue colored bars are the amount of classification gain by training the backbone with temperature variable ($\tau=0.05$).   
    }
    \label{fig:temperature_effect}
\end{figure}

We ablate the effect of having a temperature variable $\tau$ in the initial training stage using the validation set. 
As fig.~\ref{fig:temperature_effect} presents, the validation set accuracy increases with the use of $\tau$ variable across the RN-12 and RN-18. Here, ``without $\tau$'' corresponds to setting $\tau=1$, and ``with $\tau$'' to $\tau = 0.05$ (found on the validation set).

\newpage 
\section{Visualization: from MixtFSL to MixtFSL-Alignment}
\label{sec:visualization}

\begin{figure}[h!]
    \centering
    \footnotesize
    \setlength{\tabcolsep}{1pt}
    \begin{tabular}{cccc} 
    \includegraphics[height=5.5cm]{figures_plots/tsne-figures/a1_base_centroids_colored.pdf} & 
    \includegraphics[height=5.5cm]{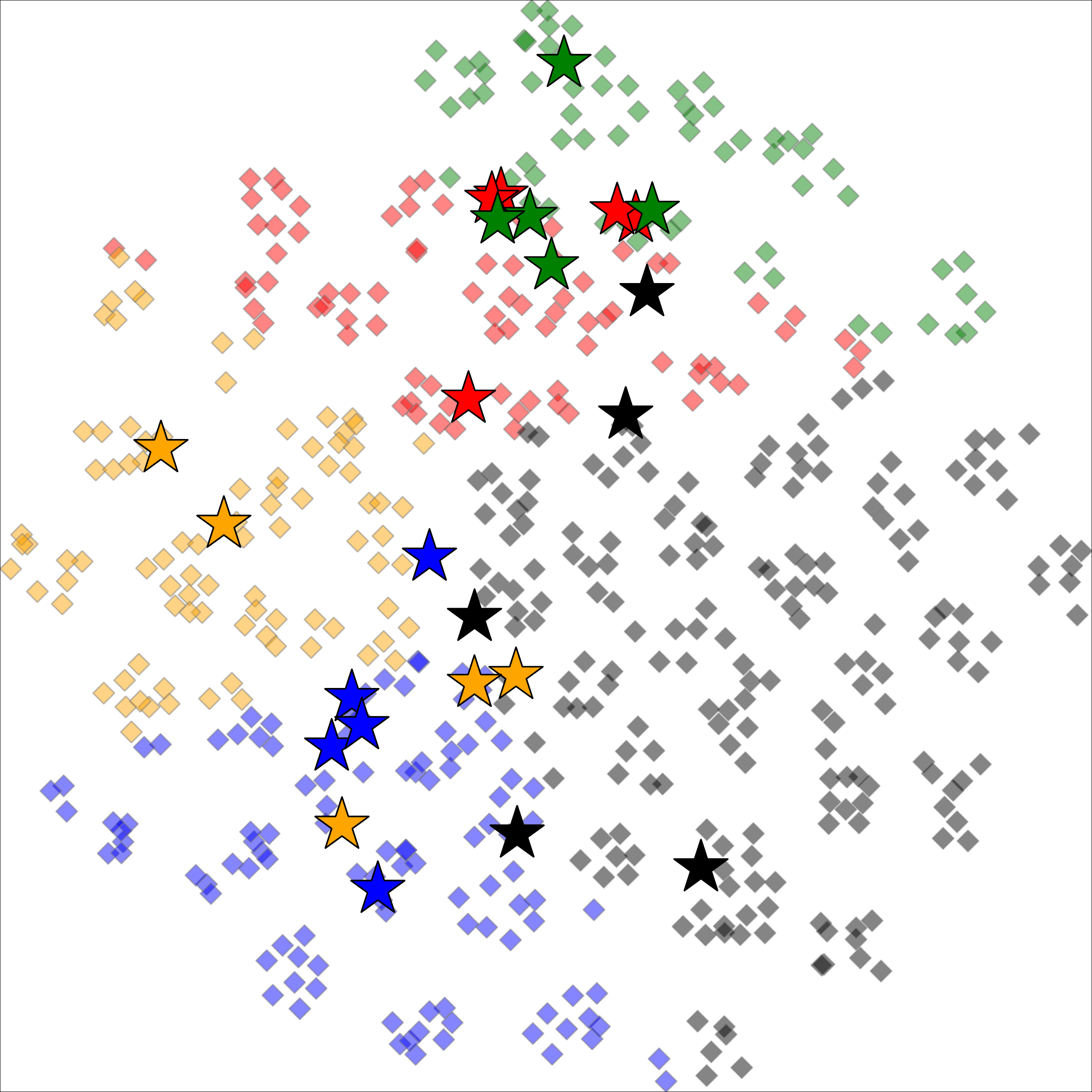} & 
    \includegraphics[height=5.5cm, angle=180,origin=c]{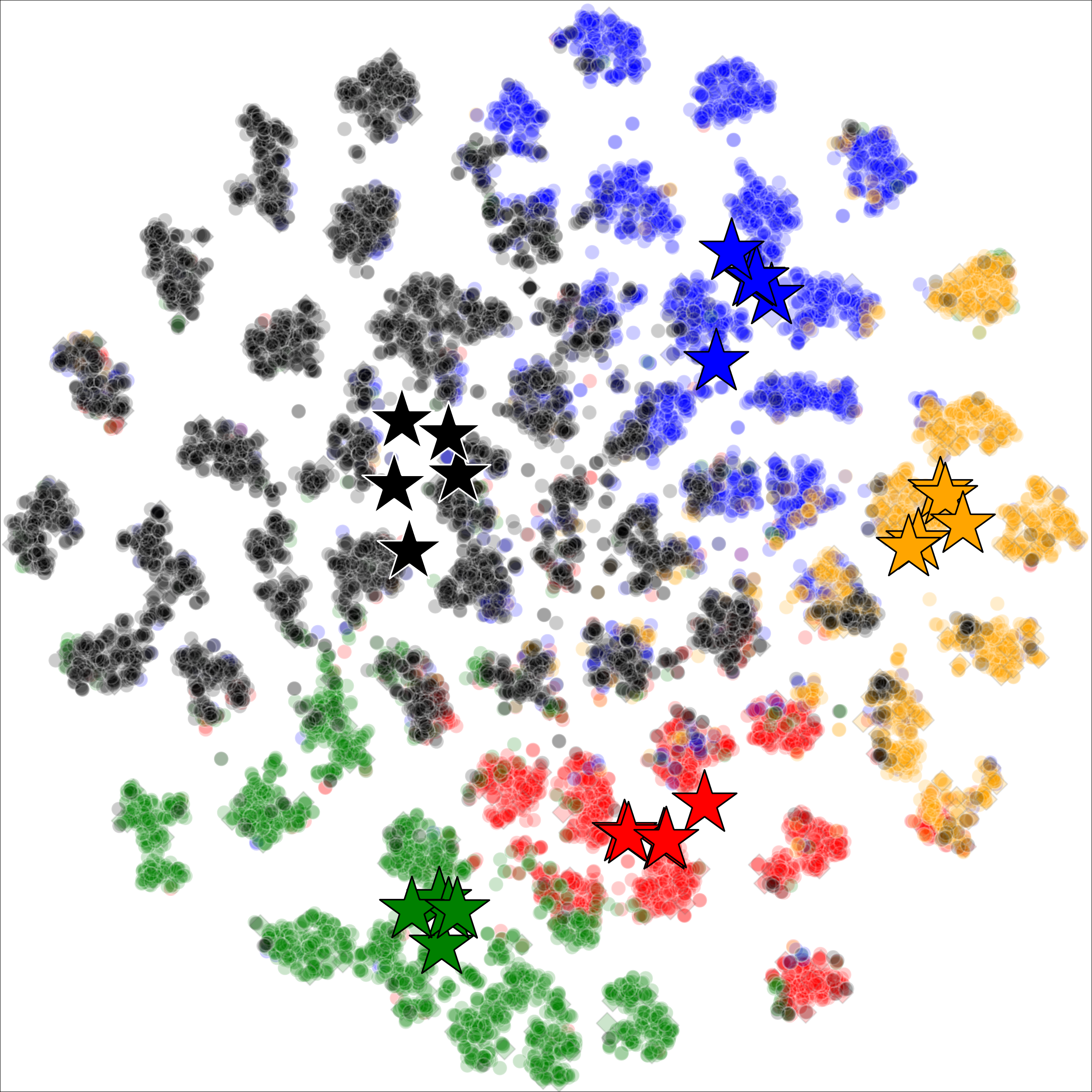} &  \\ 
    (a) after progressive following  & 
    (b) related-modes detection  & 
    (c) after alignment    &       
    \end{tabular}
    \vspace{0.5em}
    \caption{t-SNE~\cite{maaten2008visualizing} applied to the ResNet-12 base feature embedding.
    (a) learned base categories feature embedding (circles) and mixture components (diamonds) after the progressive following stages. 
    (b) using 5-way (coded by color) novel example shown by stars to detect their related base classes with the learned mixture components shown by diamonds. 
    (b) aligning the novel examples to the center of their related base classes without forgetting the base classes. Points are color-coded by related base and novel examples.  
    %\todo{Show same results but with the approach of \cite{Afrasiyabi_2020_ECCV}.}
    %\todo{Show related base samples obtained with this approach and compare to those obtained with \cite{Afrasiyabi_2020_ECCV}? }
    %\todo{Rotate to align both.}
    }
    \label{fig:Alignment}
\end{figure}

Fig.~\ref{fig:Alignment} summarizes the visualization of embedding space from our mixture-based feature space learning (MixtFSL) to its centroid alignment extension (sec.~6.1 from the main paper). Fig.~\ref{fig:Alignment}-(a) is a visualization of 200 base examples per class (circles) and the learned class mixture components (diamonds) after the progressive following training stage. Fig.~\ref{fig:Alignment}-(b) presents the t-SNE visualization of novel class examples (stars) and related base detection (diamonds of the same color) using our proposed MixtFSL. Fig.~\ref{fig:Alignment}-(c) presents the visualization of fine-tuning the centroid alignment of \cite{Afrasiyabi_2020_ECCV}. Here, the novel examples align to the center of their related bases.

% \newpage
% {\small
% \bibliographystyle{ieee_fullname}
% \bibliography{egbib}}

\end{document}